\def\year{2021}\relax
\documentclass[letterpaper]{article}
\usepackage{aaai21} 
\usepackage{times} 
\usepackage{helvet} 
\usepackage{courier} 
\usepackage[hyphens]{url} 
\usepackage{graphicx} 
\usepackage{graphicx} 
\usepackage{natbib}
\usepackage{caption}
\usepackage{subcaption}
\usepackage{algorithm}
\usepackage{algpseudocode}
\usepackage{amssymb}
\usepackage{amsmath}
\usepackage{booktabs} 
\title{Towards Trustworthy Predictions from Deep Neural Networks with Fast Adversarial Calibration}

\author{
    Christian Tomani,\textsuperscript{\rm 1,2} Florian Buettner\textsuperscript{\rm 1}\\
}
\affiliations{
    \textsuperscript{\rm 1}Siemens AG, 
    \textsuperscript{\rm 2}Technical University of Munich\\
    christian.tomani@tum.de, buettner.florian@siemens.com
}

\begin{document}
	
\maketitle

\begin{abstract} 
	To facilitate a wide-spread acceptance of AI systems guiding decision making in real-world applications, trustworthiness of deployed models is key. That is, it is crucial for predictive models to be uncertainty-aware and yield well-calibrated (and thus trustworthy) predictions for both in-domain samples as well as under domain shift. Recent efforts to account for predictive uncertainty include post-processing steps for trained neural networks, Bayesian neural networks as well as alternative non-Bayesian approaches such as ensemble approaches and evidential deep learning. Here, we propose an efficient yet general modelling approach for obtaining well-calibrated, trustworthy probabilities for samples obtained after a domain shift. We introduce a new training strategy combining an entropy-encouraging loss term with an adversarial calibration loss term and demonstrate that this results in well-calibrated and technically trustworthy predictions for a wide range of domain drifts. We comprehensively evaluate previously proposed approaches on different data modalities, a large range of data sets including sequence data, network architectures and perturbation strategies. We observe that our modelling approach substantially outperforms existing state-of-the-art approaches, yielding well-calibrated predictions under domain drift. 
\end{abstract}

\section{Introduction}
To facilitate a wide-spread acceptance of AI systems guiding decision making in real-world applications, trustworthiness of deployed models is key. Not only in safety-critical applications such as autonomous driving or medicine \citep{helldin2013presenting,caruana2015intelligible,leibig2017leveraging}, but also in dynamic open world systems in industry it is crucial for predictive models to be uncertainty-aware.  Only if predictions are calibrated in the case of any gradual domain shift, covering the entire spectrum from in-domain ("known unknowns") to truly out-of-domain samples ("unknown unknowns"), they can be trusted. 
In particular in industrial and IoT settings, deployed models may encounter erroneous and inconsistent inputs far away from the input domain throughout the life-cycle. In addition, the distribution of the input data may gradually move away from the distribution of the training data (e.g. due to wear and tear of the assets, maintenance procedures or change in customer behavior.\\ 
In recent years, a variety of different approaches to capture predictive uncertainty have been proposed. This includes intrinsically uncertainty-aware networks such as Bayesian neural networks and Deep Ensembles \cite{papernot2018deep,wen2018flipout}. Alternative approaches for obtaining meaningful predictive uncertainties rely on post-processing approaches such as Temperature Scaling \cite{guo_calibration_2017,Platt99probabilisticoutputs} \\
Efforts to evaluate the quality of predictive uncertainty have until recently been focused on in-domain calibration.\citet{guo_calibration_2017} indicate that simple post-processing methods are sufficient to yield calibrated predictions when samples are drawn from the same distribution as the training data. More recently, \citet{snoek2019can} have presented a comprehensive evaluation of calibration under domain drift for the most popular implementations of uncertainty-aware neural networks. They demonstrate that the quality of predictive  uncertainties, i.e. model calibration, decreases with increasing domain shift, regardless of method and conclude that there remains significant room for improvement.\\
In this work, we propose an efficient yet general modelling approach for obtaining well-calibrated, trustworthy probabilities under domain shift as well as for truly out-of-distribution (OOD) samples. Our approach can readily be applied to a wide range of data modalities and model architectures.  More specifically, we introduce a simple loss function to encourage high entropy on wrong predictions. We combine this with a novel adversarial calibration loss term that directly minimizes the calibration error using adversarial samples. We demonstrate that actively encouraging calibration at training time results in an uncertainty-aware neural network yielding well-calibrated predictions in the case of any gradual domain shift, from in-domain samples to truly out-of-domain samples. In contrast to the previously proposed approaches assessed in \cite{snoek2019can}, our model maintains good calibration with increasing domain shift. We illustrate this using 32 different types of dataset shift and OOD scenarios not seen during training, including the recently proposed ObjectNet benchmark dataset. Our codes are available at \url{https://github.com/tochris/falcon}. 

\section{Towards technically trustworthy predictions}\label{sec:trustworthy}

\subsection{Quantifying  calibration under domain shift: problem setup and definitions}
Let $X \in \mathbb{R}^D$ and $Y \in \{1,\dots, C\}$ be random variables that denote the $D$-dimensional input and labels in a classification task with $C$ classes, respectively. 
Let  $h(X) = (\hat{Y},\hat{P})$ be the output of a neural network classifier $h$ predicting a class $\hat{Y}$ and  associated confidence $\hat{P}$ based on $X$.\\
We are interested in the quality of the predictive uncertainty of $h$ (i.e. confidence scores  $\hat{P}$) and quantify this quality using the notion of \it calibration \rm. In particular, we are interested in the calibration of neural networks under dataset shift. That is, we assess calibration of neural networks for test data, where the distribution of samples seen by a model gradually moves away from the training distribution (in an unknown fashion) until it reaches truly OOD levels.\\ 
We follow \citet{guo_calibration_2017} and formally define perfect calibration such that confidence and accuracy match for all confidence levels:
$$\mathop{\mathbb{P}}(\hat{Y}=Y| \hat{P}=p) = p, \; \; \forall p \in [0,1]$$ 
This directly leads to a definition of miss-calibration as the difference in expectation between confidence and accuracy, that is $ \mathop{\mathbb{E}}_{\hat{P}}\left[\big\lvert\mathop{\mathbb{P}}(\hat{Y}=Y| \hat{P}=p) - p\big\rvert \right]$. Miss-calibration can be estimated from finite samples by  partitioning  predictions  into $M$ equally-spaced bins and computing a weighted average of the bins' difference between accuracy and confidence. The resulting measure is the  expected calibration error (ECE) \cite{naeini2015obtaining}:
\begin{eqnarray}
    \mathrm{ECE} = \sum_{m=1}^M \frac{\lvert B_m\rvert}{n}\big\lvert \mathrm{acc}(B_m) - \mathrm{conf}(B_m)\big\rvert
\end{eqnarray}\label{eq:ece}
with $B_m$ being the set of indices of samples whose prediction confidence falls into its associated interval $I_m$. $\mathrm{conf}(B_m)$ and $\mathrm{acc}(B_m)$ are the average confidence and accuracy associated to $B_m$ respectively and $n$ the number of samples in the dataset.

\subsection{Related work}
\begin{figure*}[t]
	\begin{minipage}{\textwidth}
		\begin{algorithm}[H]
			\caption{FALCON with set of perturbation levels \\ $\mathcal{E}=\{0, 0.05, 0.1, 0.15, 0.2, 0.25, 0.3, 0.35, 0.4, 0.45\}$ , mini batch size $b$, and training set $(X, Y)$.  
			}\label{alg1}
			\begin{algorithmic}[1]
				\Repeat
				\State Read minibatch $MB = (\{X_1 ,\dots, X_b\}, \{Y_1 , \dots, Y_b\})$
				\State Randomly sample $\epsilon_{MB}$ from $\mathcal{E}$
				\State Generate FGSM minibatch $MB_{adv}$  of size $b$ from samples in $MB$ using $\epsilon_{MB}$
				\State Compute $L_{CCE}$ and $L_S$ and do one training step using mini batch $MB$
				\State Compute predictions for all samples in $MB_{adv}$ and partition into $M$ equally spaced bins
				\State Compute binned accuracy $\mathrm{acc}(B_{m_i})$ for all samples $i$ in $MB_{adv}$
				\State Compute $L_{adv}$ based on $MB_{adv}$ and do one training step using $MB_{adv}$
				\Until{training converged}
			\end{algorithmic}
		\end{algorithm}
	\end{minipage}
\end{figure*}
Related work can be broadly divided in 2 groups, namely intrinsically uncertainty-aware neural networks and post-processing methods for a post-hoc transformation of uncalibrated confidence scores.\\
Different avenues towards intrinsically uncertainty-aware networks exist. In particular, a lot of research effort has been put into training probabilistic models such as Bayesian neural networks. For these models, typically a prior distribution over the weights is specified and, given the training data, a posterior distribution over the weights is inferred. This distribution can then be used to quantify predictive uncertainty. Since exact inference is intractable, a range of approaches for approximate inference has been proposed. In particular approaches based on variational approximations have recently received a lot of attention \cite{blundell2015weight} and range from the interpretation of Gaussian dropout as performing approximate Bayesian inference  \cite{gal2016dropout} to facilitating a complex posterior using normalising flows \cite{louizos_multiplicative_2017} and stochastic variational inference based on Flipout \cite{wen2018flipout}. Since such Bayesian approaches often come at a high computational cost, alternative non-Bayesian approaches have been proposed, that can also account for predictive uncertainty. These include ensemble approaches, where  smooth predictive estimates can be obtained by training ensembles of neural networks using adversarial examples \cite{lakshminarayanan_simple_2017}, and evidential deep learning, where predictions of a neural net are modelled as subjective opinions by placing a Dirichlet distribution on the class probabilities \cite{sensoy_evidential_2018}. More recently, \citet{thulasidasan2019mixup} have shown that using MixUp training, where label- and input smoothing is performed, yields good results for in-domain calibration.\\
An alternative strategy towards uncertainty-aware neural networks is based on post-processing steps for trained neural networks. For these methods, a validation set, drawn from the same distribution as the training data, is used to post-process the logit vectors returned by a trained neural network such that in-domain predictions are well calibrated. The most popular approaches include parametric methods such as Temperature Scaling and Platt Scaling as well as non-parameteric methods such as isotonic regression and histogram binning  \cite{Platt99probabilisticoutputs,guo_calibration_2017}. More recently, \citet{kumar2019verified} have proposed an alternative approach for post-processing combining histogram binning with a parametric postprocessing function and demonstrated that this scaling-binning calibrator results in better in-domain calibration than Temperature scaling.\\
Orthogonal approaches have been proposed where trust scores and other measures for OOD detection are derived, typically also based on trained networks \cite{liang_enhancing_2018,jiang2018trust,papernot2018deep}; however these latter approaches differ substantially in their modeling assumptions from the models described above. While intrinsically uncertainty-aware neural networks primarily differ from each other in how they quantify predictive uncertainty $\mathop{\mathbb{P}}(Y|X)$, many OOD methods introduce an additional component to $\mathop{\mathbb{P}}(Y|X)$ such as trust scores, which are inherently different from probabilities $\mathop{\mathbb{P}}(Y|X)$, and cannot readily be compared to predictive entropy. Other methods need access to known OOD datasets during training or train GANs in addition to a classifier \cite{lee2017training,hendrycks2018deep}. \citet{snoek2019can} argue that it is difficult to perform a meaningful comparison between such OOD methods and intrinsically uncertainty-aware networks. We follow their setup and focus on intrinsically uncertainty-aware networks and post-processing approaches, which make the same assumptions about data, in this study.\\
Finally, intrinsically uncertainty-aware neural networks is a very active research field and independent and concurrent studies include new takes on Bayesian neural networks \cite{joo2020being,chan2020unlabelled} and an extension of MixUp \cite{zhang2020mix}.

\section{Fast adversarial calibration}

\subsection{Predictive entropy loss}
Here, we propose a new, simple approach based on fast adversarial calibration to obtain well-calibrated uncertainty estimates for predictions under domain shift. To mitigate overconfident predictions displayed by conventional deep neural networks \cite{guo_calibration_2017,snoek2019can}, we first introduce a cross-entropy loss term encouraging a uniform distribution of the scores in case the model "does not know". That is, for each sample we remove the confidence score corresponding to the correct label and compute the cross-entropy between a uniform distribution and the remaining false confidence scores in order to distribute the probability mass of these false predictions uniformly over $C$ classes:
\begin{eqnarray}
	L_S = \sum_{i=1}^{n} \sum_{j=1}^{C}-\frac{1}{C}\log(p_{ij}(1-y_{ij}) + y_{ij}), 
\end{eqnarray}
$p_{ij}$ denotes the confidence associated to the $j$th class of sample $i$, $y_{ij}$ its one-hot encoded label.\\
This simple loss term increases uncertainty-awareness by encouraging an increased entropy ($S$) in the presence of high predictive uncertainty without directly affecting reconstruction loss (categorical cross-entropy). This has the advantage that our approach - in contrast to state-of-the-art Bayesian neural networks such as those based on multiplicative normalizing flows or evidential deep learning - can be readily applied to complex architectures based on LSTMs or GRUs. 

\subsubsection{Adversarial calibration loss}
\begin{figure*}[t]
	\centering 
	\begin{subfigure}[t]{0.31\textwidth}
		\centering
		\includegraphics[width=\textwidth]{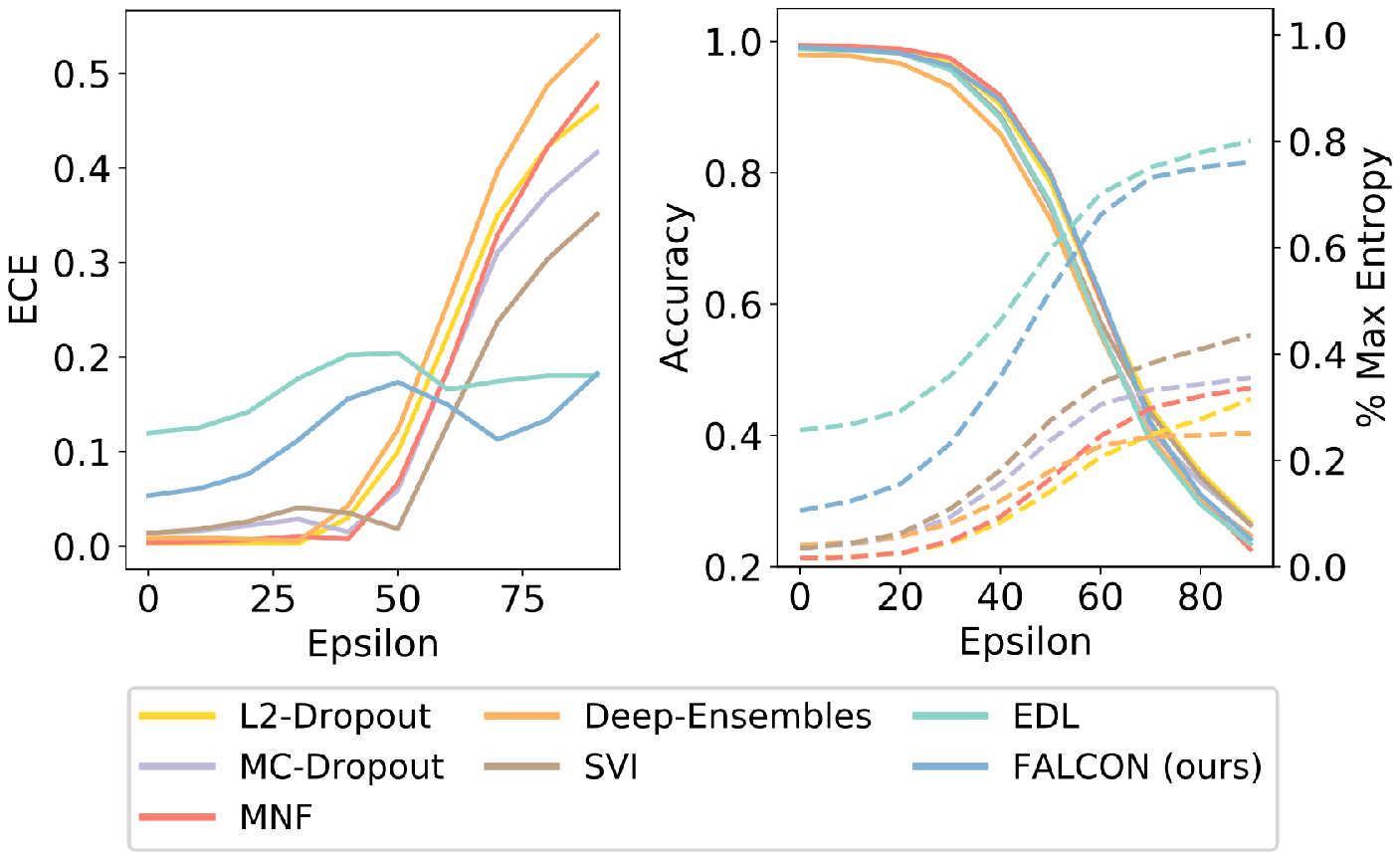}
		\caption{Calibration under y-zoom}
	\end{subfigure}
	\begin{subfigure}[t]{0.45\textwidth}
		\centering
		\includegraphics[width=\textwidth]{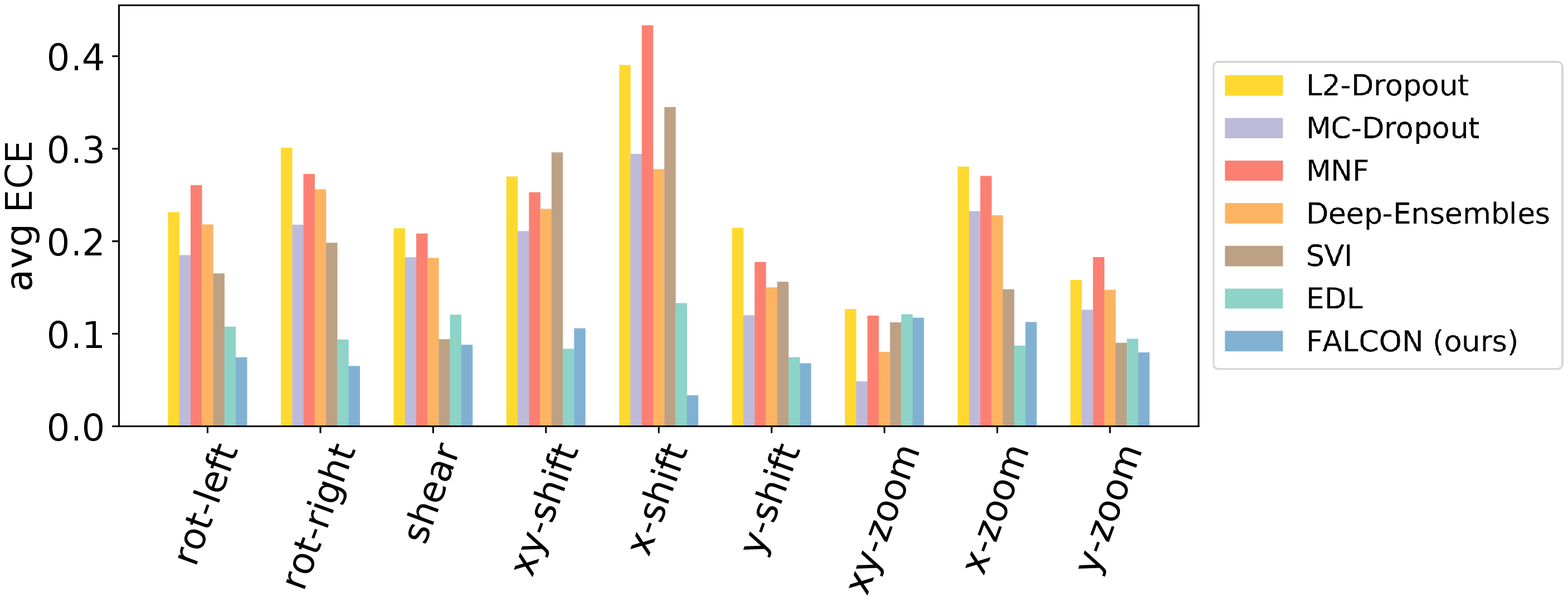}
		\caption{ECE for all perturbations}
    \end{subfigure}
	\caption{Calibration of the predictive uncertainty under domain shift MNIST data.  (a) \textit{Left}: ECE at 10 increasingly large levels of y-zoom. Only EDL and FALCON maintain a low ECE across all levels of y-zoom. \textit{Middle:} Entropy increases with larger y-zoom for all methods. While EDL starts at the highest entropy, this reflects under-confident predictions for low levels of perturbation. (b) FALCON results in consistently well calibrated and robust predictions across all perturbation strategies.}
	\label{fig:accS}
\end{figure*}
\begin{figure*}
	\centering
	\begin{subfigure}[t]{0.19\textwidth}
		\centering
		\includegraphics[width=\textwidth]{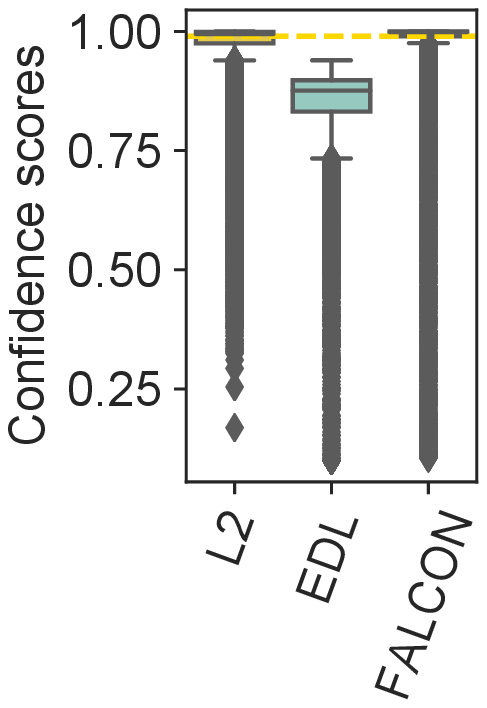}
		\caption{In-domain calibration}
	\end{subfigure}
	\begin{subfigure}[t]{0.3\textwidth}
		\centering
		\includegraphics[width=\textwidth]{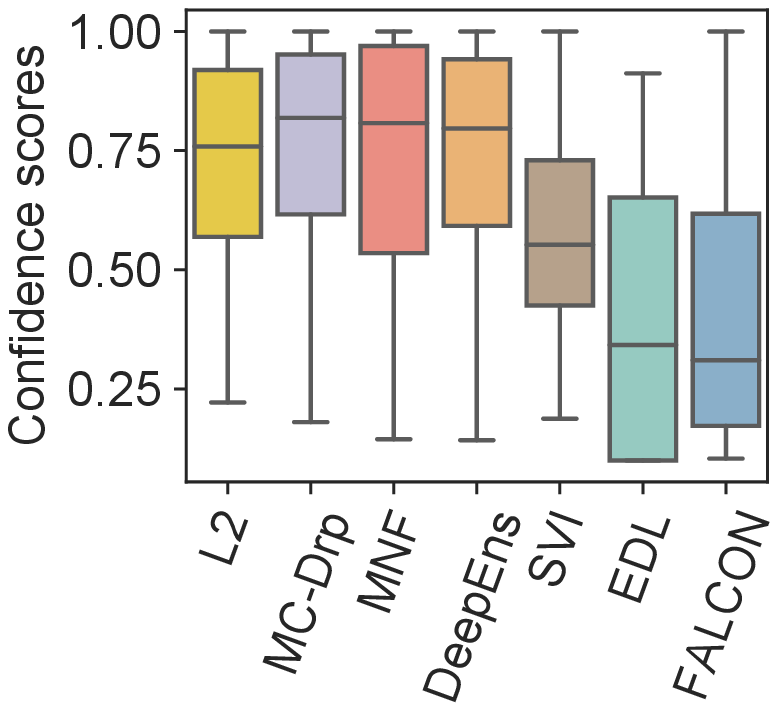}
		\caption{OOD calibration MNIST}
	\end{subfigure}
	\begin{subfigure}[t]{0.49\textwidth}
		\centering
		\includegraphics[width=\textwidth]{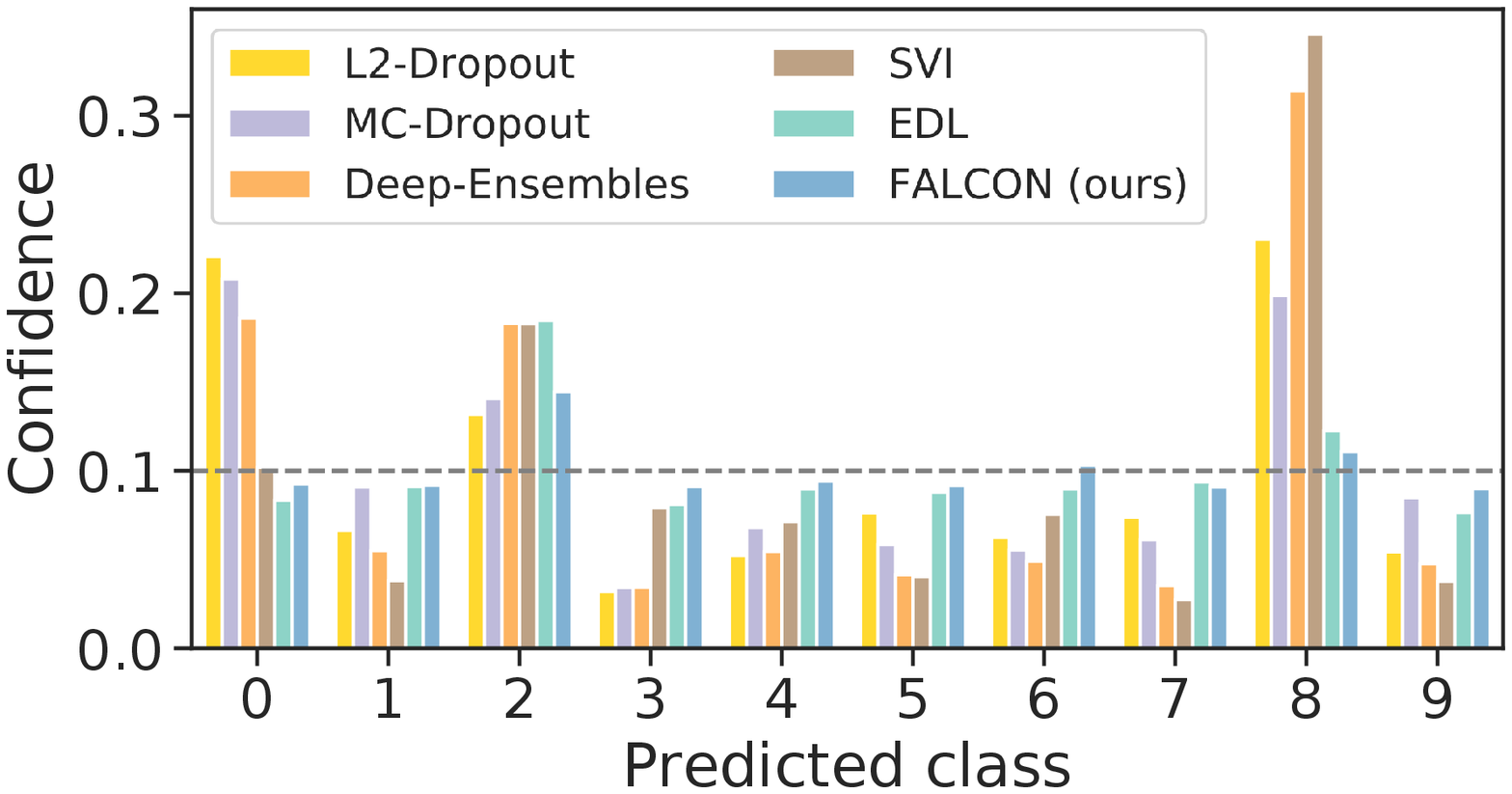}
		\caption{OOD calibration Fashion-MNIST}
	\end{subfigure}
	\caption{Distribution of confidence scores for in-domain and truly OOD scenarios. (a) Confidence scores of L2 and FALCON match accuracy (dashed yellow line) on MNIST, whereas EDL is significantly more underconfident than FALCON ($p<10^{-10}$, Wilcoxon signed rank test). (b) Confidence scores of all truly OOD scenarios (perturbations with close-to-random accuracy ($<.2$) at epsilon 90) for MNIST. (c) Average confidence per predicted class for Fashion MNIST data should be close to random for all classes; this is only the case for FALCON. 
	}
	\label{fig:box_conf}
\end{figure*}
While the entropy-based loss term does encourage uncertainty-awareness, we found that it is beneficial to introduce an additional loss term addressing model calibration directly. Explicitly encouraging calibration for out-of-domain samples, however - e.g. via an ECE-based measure - requires knowledge on the type of perturbed or erroneous samples the model is expected to encounter. In many real-world applications it is not clear from which distribution these samples will be drawn and, for model predictions to be truly trustworthy requires robustness against all such potential out-of-domain samples. That is, we would like our model to be technically robust for inputs around an $\epsilon$-neighbourhood of the in-domain training samples, for a wide range of $\epsilon$ and for all $2^D$ directions in $\{-1, 1\}^D$. While inputs from a random direction are unlikely to be representative examples for generic out-of-domain samples, by definition adversarial examples are generated along a dimension where the loss is high. \citet{lakshminarayanan_simple_2017} show that adversarial training can improve the smoothness of predictions, in particular when training an ensemble of 5 neural networks in an adversarial fashion. Here, we demonstrate that using adversarial samples to directly optimise model calibration (rather than the squared error of one-hot encoded labels \cite{lakshminarayanan_simple_2017}) results in substantially more trustworthy predictions for out-of-domain samples from a large number of unrelated directions.\\
We implement an ECE-inspired calibration loss by minimizing miss-calibration for samples generated using the fast gradient sign method (FGSM) \cite{goodfellow2014explaining}, with $\epsilon$ ranging from 0 to 0.5 (sampled at 10 equally spaced bins at random). To this end we minimise the L2 norm of the difference between the predicted confidence of a sample $i$, which we denote as $\mathrm{conf}(i)$, and its corresponding binned accuracy $\mathrm{acc}(B_{m_i})$, for all samples. This is directly motivated by the definition of ECE (eq. 1), except that we re-write eq. 1 by explicitly summing over all samples and replace the L1 norm with the L2 norm. As for the computation of ECE, we partition the predictions of the network into $M$ equally-spaced bins, with $m_i \in \{1,\dots,M\}$ being the bin into which sample $i$ falls. As for ECE, $B_{m_i}$ is the set of indices of samples falling in bin $m_i$ and $\mathrm{acc}(B_{m_i})$ the average accuracy of samples $B_{m_i}$.
We set $M=10$ for all experiments.
\begin{eqnarray}
	L_{\textrm{adv}} &=& \sqrt{\sum_{i=1}^n \left(\mathrm{acc}(B_{m_i}) - \mathrm{conf}(i)\right)^2} 
\end{eqnarray}
The final loss balancing a standard reconstruction loss (categorical cross entropy (CCE)) against the entropy and adversarial calibration loss  can then be written as $
L = L_{\textrm{CCE}} + \lambda_{\textrm{adv}}  L_{\textrm{adv}} + \lambda_S L_S$. The choice of hyperparameters $\lambda_{\textrm{adv}}$ and $\lambda_S$ is described in  the supplement. Note that we do not use the FGSM samples for adversarial training in the sense that we do not try to minimize the reconstruction error (cross entropy) for those samples. The algorithm is summarized in Algorithm 1. We refer to our method based on Fast AdversariaL CalibratiON as FALCON.
\section{Experiments and results}
\begin{figure*}
	\centering
	\begin{subfigure}[t]{0.57\textwidth}
		\centering
		\includegraphics[width=\textwidth]{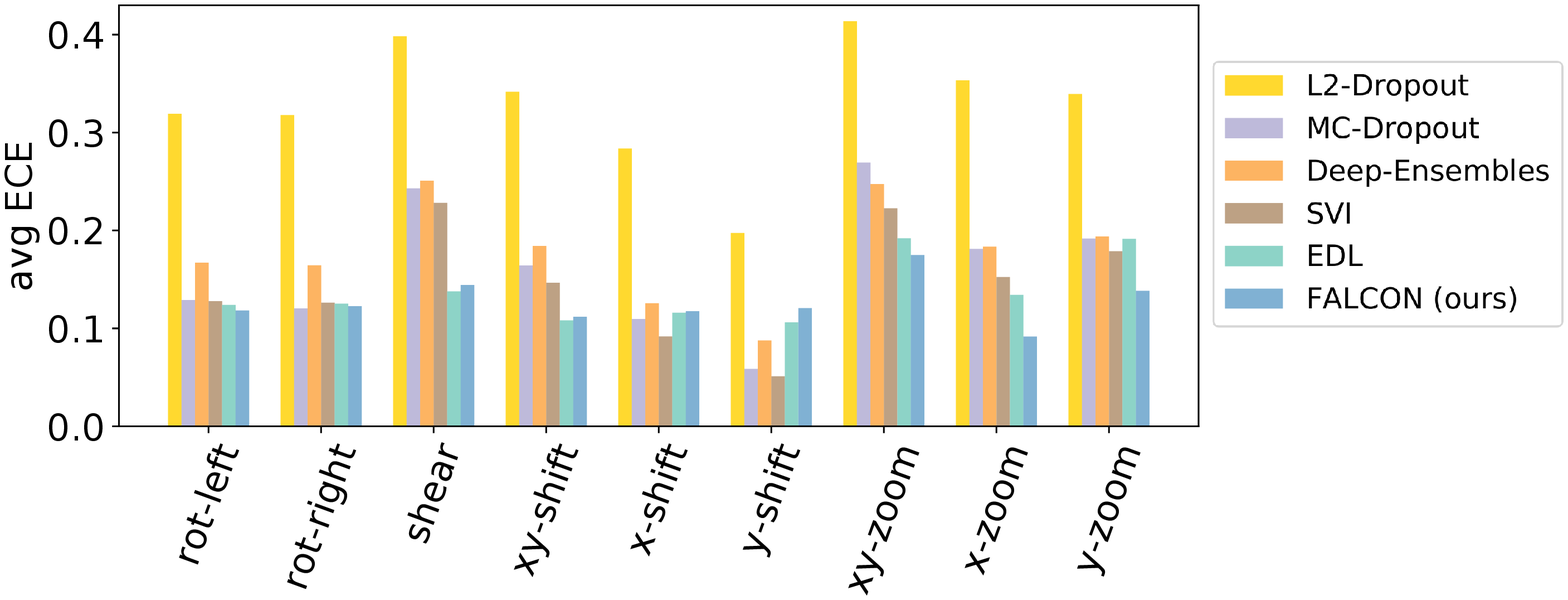}
		\caption{Calibration under domain shift}
	\end{subfigure}
	\begin{subfigure}[t]{0.25\textwidth}
		\centering
		\includegraphics[width=\textwidth]{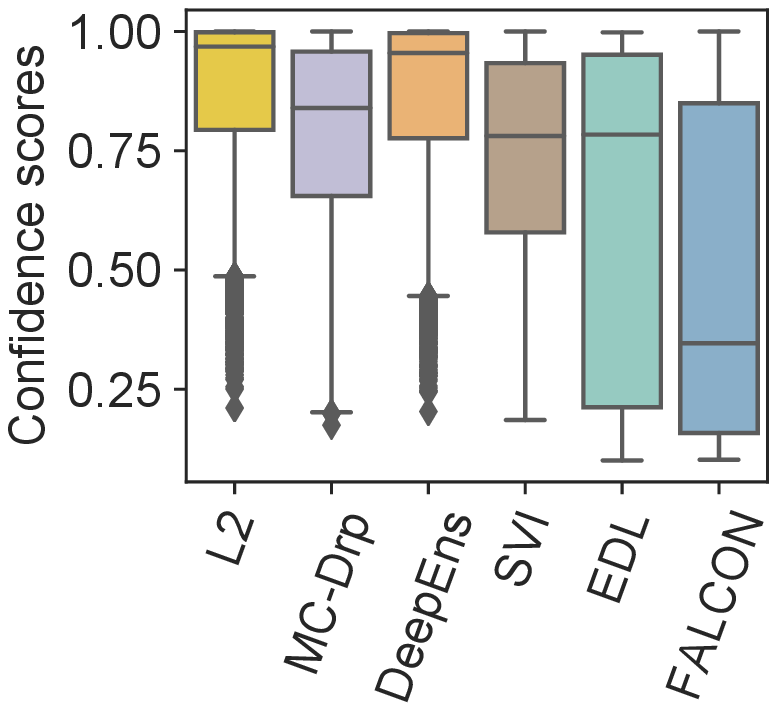}
		\caption{Predictive uncertainty for truly OOD samples}
	\end{subfigure}
	\caption{Calibration of image classification models, quantified by computing the micro-averaged expected calibration error (lower is better). FALCON results in consistently well calibrated and robust predictions across all perturbation strategies.}
	\label{fig:cifar}
\end{figure*}
\begin{figure}
    \centering
    \includegraphics[width=1\columnwidth]{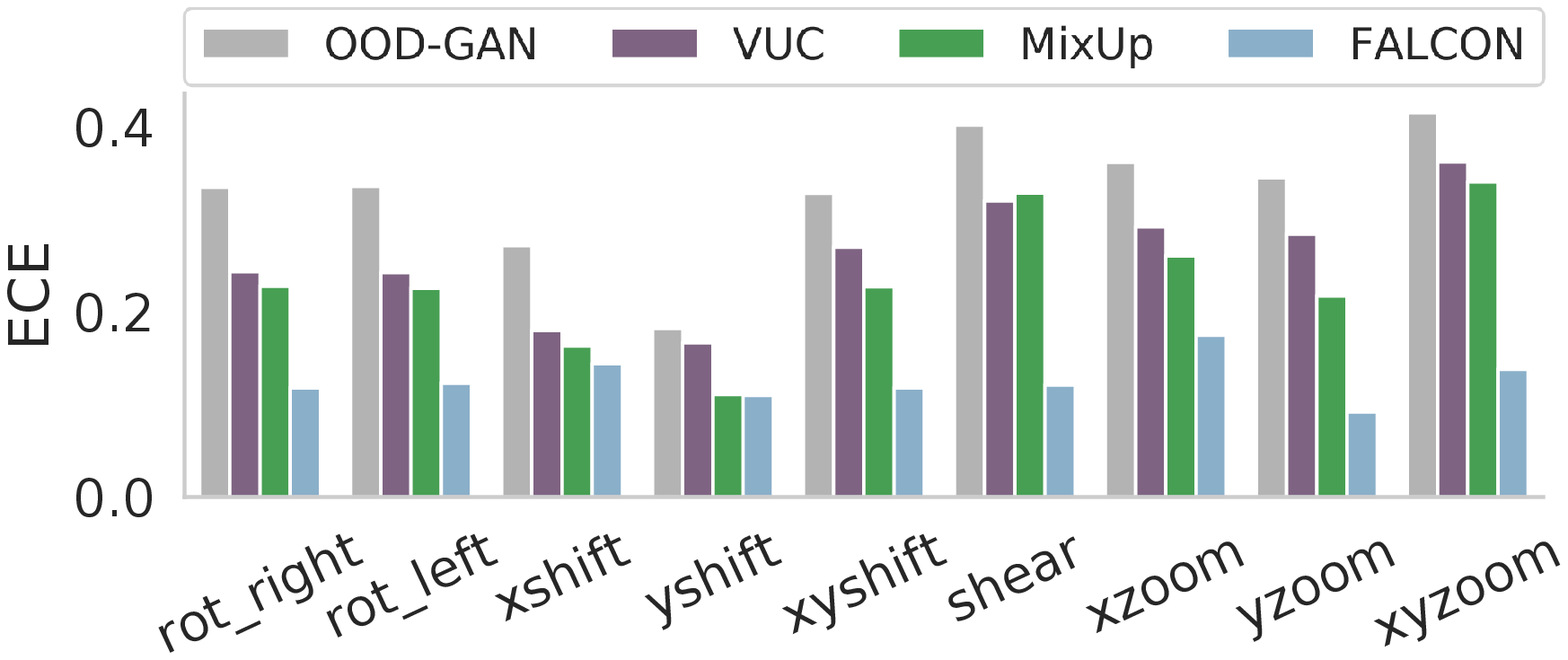}
    \caption{Micro-averaged expected calibration error (lower is better) for CIFAR10 for additional baselines. FALCON performs consistently better than those baselines, too.} \label{fig:newbase}
\end{figure}

We quantify calibration under domain shift for 29 distinct perturbation types not seen during training, including affine transformations, image corruptions and word swaps as well as a dedicated bias-controlled dataset \cite{barbu2019objectnet}. Each perturbation strategy mimics a scenario where the data a deployed model encounters stems from a distribution that gradually shifts away from the training distribution in a different manner.  For each model and each perturbation, we compute the micro-averaged ECE by first perturbing each sample in the test set at 10 different levels and then calculating the overall ECE across all samples; we denote relative perturbation strength as epsilon (see Appendix for mapping to absolute values).\\
In addition, we quantify the quality of predictive uncertainty for truly OOD scenarios by computing the predictive entropy and distribution of confidence scores. We use complete OOD datasets as well as data perturbed at the highest level. In these scenarios we expect entropy to reach maximum levels, since the model should transparently communicate it "does not know" via low and unbiased confidence scores.\\ 

We show that our modelling approach substantially outperforms existing approaches for sequence models and illustrate improved performance for image data. 

We compared the following modelling approaches: (i) \textit{L2-Dropout}, referring to  a standard neural net with L2 regularisation and dropout as baseline, (ii) \textit{MC-Dropout} corresponding to the modelling approach presented by \cite{gal2016dropout}, (iii) \textit{Deep Ensembles} referring to an approach based on an ensemble of neural nets \cite{lakshminarayanan_simple_2017}, (iv) \textit{EDL} referring to Evidential Deep Learning \cite{sensoy_evidential_2018}, (v) \textit{MNF} referring to a Bayesian neural network trained using multiplicative normalising flows \cite{louizos_multiplicative_2017}, (vi) \textit{SVI}, referring to stochastic variational inference based on Flipout \cite{wen2018flipout}, (vii) \textit{FALCON}, which is our proposed method. We also compare the additional baselines verified uncertainty calibration (VUC) \cite{kumar2019verified}, and MixUp \cite{thulasidasan2019mixup}; to illustrate that the different modeling assumptions of OOD detection methods do not translate into calibrated predicted uncertainty under domain drift, we also jointly trained a classifier and a GAN \cite{lee2017training}. Since \citet{snoek2019can} have comprehensively shown that Temperature Scaling results in substantially higher overconfidence than other baselines, we only report results on this baseline in the Appendix and refer to \cite{snoek2019can} for a more detailed analysis.
Additional information on model training, parameter and hyperparameter settings for all methods (including ablation studies and sensitivity analyses for FALCON) as well as perturbation strategies is given in the Appendix. 
\begin{table}
	\centering
	\caption{Test accuracy and mean ECE across all 9 perturbation strategies for MNIST and CIFAR10.}
	\begin{tabular}{l l l l l}
		\toprule
		&  \multicolumn{2}{c}{MNIST} & \multicolumn{2}{c}{CIFAR10}\\
		& Acc. & ECE & Acc. & ECE  \\ 
		\midrule
		L2-Drp & 0.99 & 0.243 & 0.872 & 0.329 \\ 
		MC-Drp & 0.992 & 0.179 & 0.865 & 0.163 \\ 
		MNF & 0.993 & 0.197 & NA & NA \\ 
		DeepEns & 0.98 & 0.242 & 0.868 & 0.178 \\ 
		SVI & 0.989 & 0.184 & 0.867 & 0.147 \\
		EDL & 0.989 & 0.102 & 0.87 & 0.137 \\ 
		MixUp & 0.991 & 0.20 & 0.871 & 0.23 \\ 
		VUC & 0.99 & 0.45 & 0.872 & 0.26 \\ 
		FALCON & 0.991 & \textbf{0.082} & 0.866 & \textbf{0.127} \\ 
		\bottomrule
	\end{tabular}
	\label{tab:ece_im}
\end{table}
\subsection{Predictive uncertainty for image classification}
\paragraph{MNIST} We first trained the existing baseline approaches and evaluated them on 9 different perturbation strategies (not used during training) on MNIST. While with increasingly strong perturbations the predictive entropy increased for all models, this was not necessarily matched by a good calibration across the range of the perturbation. At the typical example of the perturbation y-zoom, it becomes clear that for most methods entropy did not increase sufficiently fast to match the decrease in accuracy, resulting in increasingly overconfident predictions and an increasing ECE for stronger perturbations (Fig. \ref{fig:accS} (a)).\\
We observed a similar behaviour across all other 8 perturbation strategies, which was reflected in the lowest micro-averaged ECE for FALCON, followed by EDL (Figure \ref{fig:accS} (b); Table \ref{tab:ece_im}).\\
\textit{In-domain and truly OOD calibration}  While FALCON and EDL yielded well-calibrated predictions that were robust across all perturbation levels, it is worth noting that EDL has a substantially higher ECE for in-domain predictions, reflecting under-confident predictions on the test set (Fig. \ref{fig:box_conf} (a)). 

\begin{figure}
	\centering
	\includegraphics[width=\columnwidth]{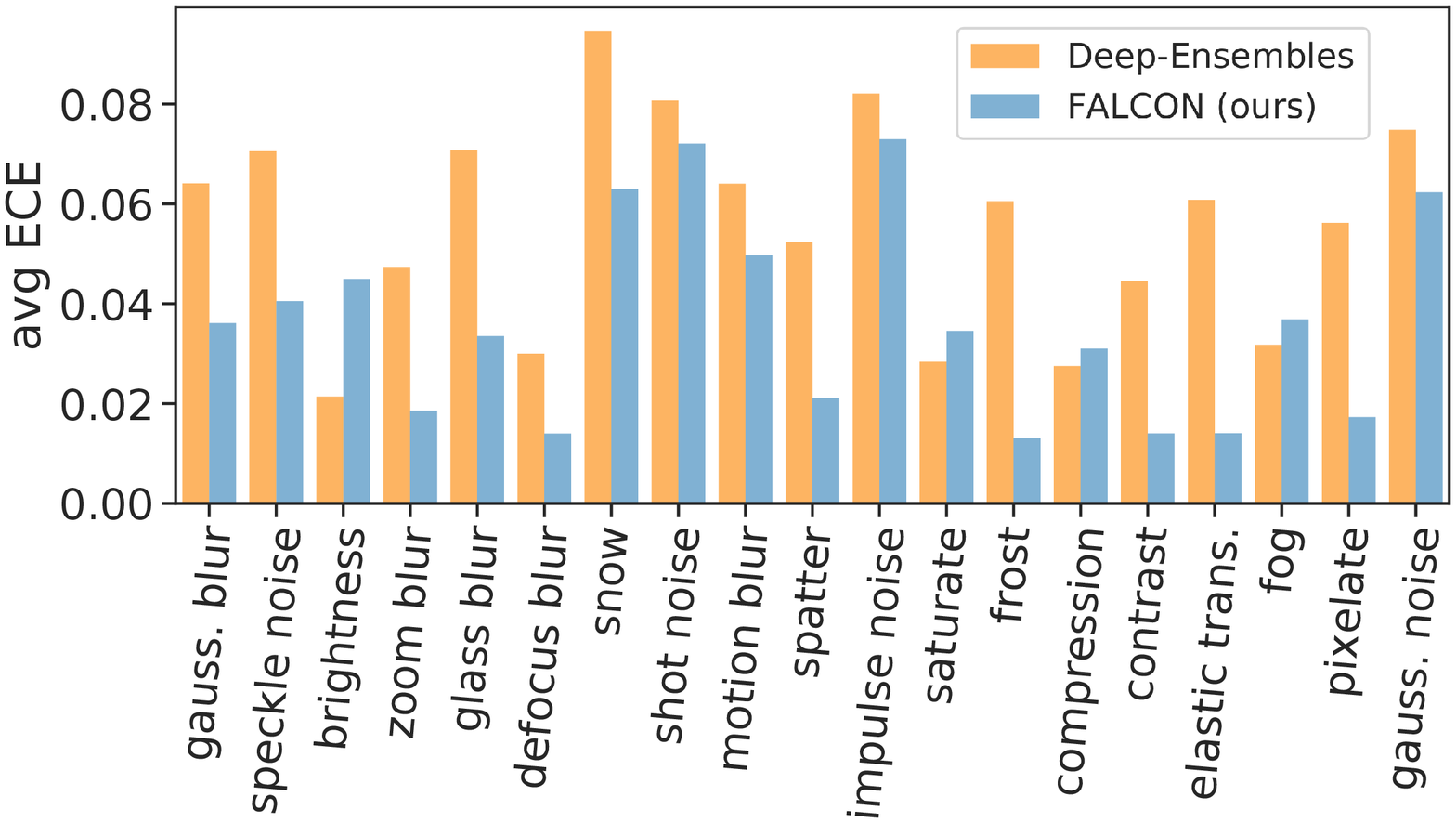}
	\caption{ Micro-averaged ECE for 19 ImageNet corruptions.}
	\label{fig:imagenetcorr}
\end{figure}

We further assessed the quality of the predictive uncertainties for truly OOD samples using the complete OOD dataset Fashion MNIST \cite{xiao2017fashion}. For such OOD samples, models are not able to make meaningful predictions, which should be reflected  in consistently low confidence scores for all predicted classes. Figure \ref{fig:box_conf} (b) shows the mean confidence scores per predicted class. This illustrates the highly overconfident OOD predictions made by baseline methods and their undesired bias towards a subset of classes. Note that detailed results for the additional baselines, including OOD performance, are reported in the Appendix.
\paragraph{CIFAR10} Next, we trained a VGG19 model on the CIFAR10 dataset. We again observed a similar trend as for the MNIST data, with FALCON yielding well calibrated predictions across all perturbation  strategies (Fig.  \ref{fig:cifar}). The considerable overconfidence of most baseline methods when making predictions on OOD samples is also reflected by the distribution of OOD confidence scores (Fig. \ref{fig:cifar}). Considering perturbations at maximum strength (epsilon 90), FALCON is the only model yielding uncertainty-aware confidence scores at a median of less than 0.5 (for VUC and MixUp see Fig. \ref{fig:newbase}).\\
We omitted MNF due to the large memory requirements stemming from the use of multiplicative normalising flows.
\begin{figure}
	\centering
	\begin{subfigure}[t]{0.125\textwidth}
		\centering
		\includegraphics[width=\textwidth]{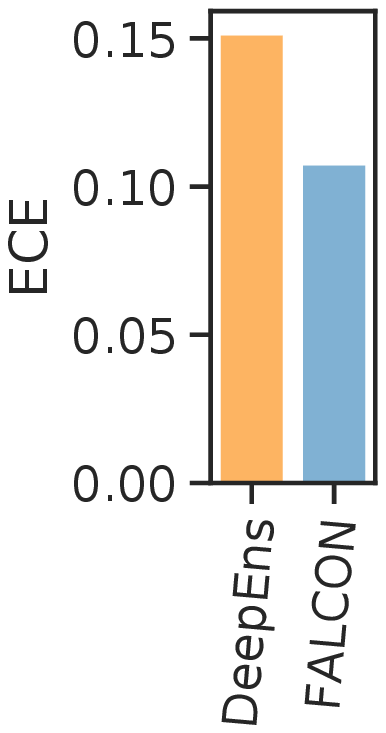}
		\caption{Overlap}
	\end{subfigure}
	\begin{subfigure}[t]{0.1\textwidth}
		\centering
		\includegraphics[width=\textwidth]{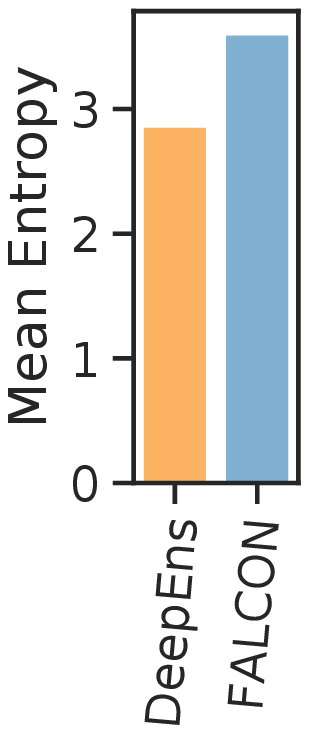}
		\caption{OOD}
	\end{subfigure}
	\begin{subfigure}[t]{0.12\textwidth}
		\centering
		\includegraphics[width=\textwidth]{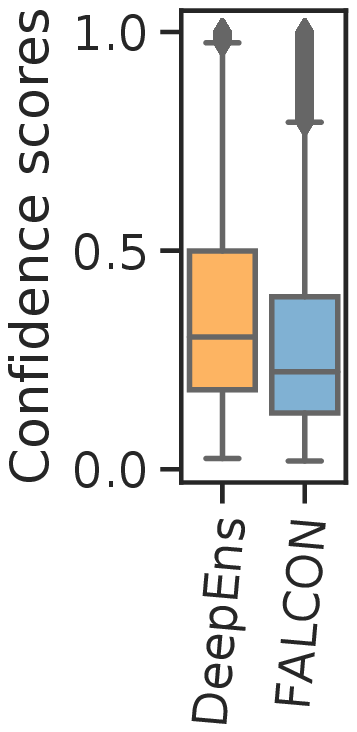}
		\caption{OOD}
	\end{subfigure}	
	\caption{Calibration of ImageNet models for domain shift and OOD scenarions (a) ECE for ObjectNet overlapping classes (lower is better), (b) predictive entropy for ObjectNet OOD classes (higher is better) and (c) confidence scores for ObjectNet OOD classes (lower is better).}
	\label{fig:imagenet}
\end{figure}	
\paragraph{ImageNet} 
To evaluate the technical robustness and calibration of FALCON on large-scale data and a more complex architecture, we quantified the expected calibration error for Resnet50 models trained on ImageNet.  \citet{snoek2019can} have shown that Deep Ensembles clearly outperform all other baselines on ImageNet in terms of calibration. We therefore use the trained Deep Ensemble consisting of 10 neural networks provided as part of \cite{snoek2019can} and compare FALCON to this strongest baseline.\\
A common manifestation of dataset shift in real-world applications is a change in object backgrounds, rotations, and imaging viewpoints. In order to quantify the expected calibration error under those scenarios, we use ObjectNet, a recently proposed large-scale bias-controlled dataset \cite{barbu2019objectnet}. In addition, we compute the ECE on various unseen test perturbations by investigating a set of 19 recently introduced corruptions \cite{hendrycks2019benchmarking}.\\
The ObjectNet dataset contains 50,000 test images with a total of 313 classes, of which 113 overlap with ImageNet. We first compute accuracy and ECE for the overlapping classes and use the non-overlapping classes as OOD test set.
While accuracy is comparable between FALCON and Deep Ensembles with 0.754 and 0.783 respectively, Deep Ensembles benefit the ensemble effect, which typically results in improved model fits. FALCON is also able to benefit from this effect and an ensemble of FALCON results in the same accuracy as deep ensembles (0.782); this effect also holds for the negative log likelihood and under domain shift (see Appendix for in-depth analysis).
As previously reported, removing typical biases (e.g. correlation between background and object class) results in an approximately 40\% decrease in accuracy for all models (\cite{barbu2019objectnet}). While ultimately generalization of models to such de-biased datasets is desirable, for many real world applications it is critical that models are transparent in settings where such generalization is not achieved.
Computing the expected calibration error reveals that FALCON is substantially more transparent about this drop in accuracy, with considerably lower ECE than Deep Ensembles (Fig. \ref{fig:imagenet} a). Similarly, we find that FALCON outperforms Deep Ensembles also for 19 image corruptions, with the mean ECE across all perturbations being 35\% lower than for the Deep Ensemble (0.056 vs. 0.036; Fig. \ref{fig:imagenetcorr}).\\
To investigate OOD calibration of ImageNet models, we use the 200 classes in Objectnet that do not overlap with ImageNet classes as completely OOD dataset. FALCON had the most uncertainty-aware predictions, with predictive entropy being higher than for the Deep Ensemble and confidence scores being significantly lower ($p<10^{-10}$, Wilcoxon signed rank test; Fig. \ref{fig:imagenet} b and c).
\subsection{Predictive uncertainty for sequence modeling}
\begin{figure}
	\centering
	\begin{subfigure}[t]{0.72\columnwidth}
		\centering
		\includegraphics[width=\textwidth]{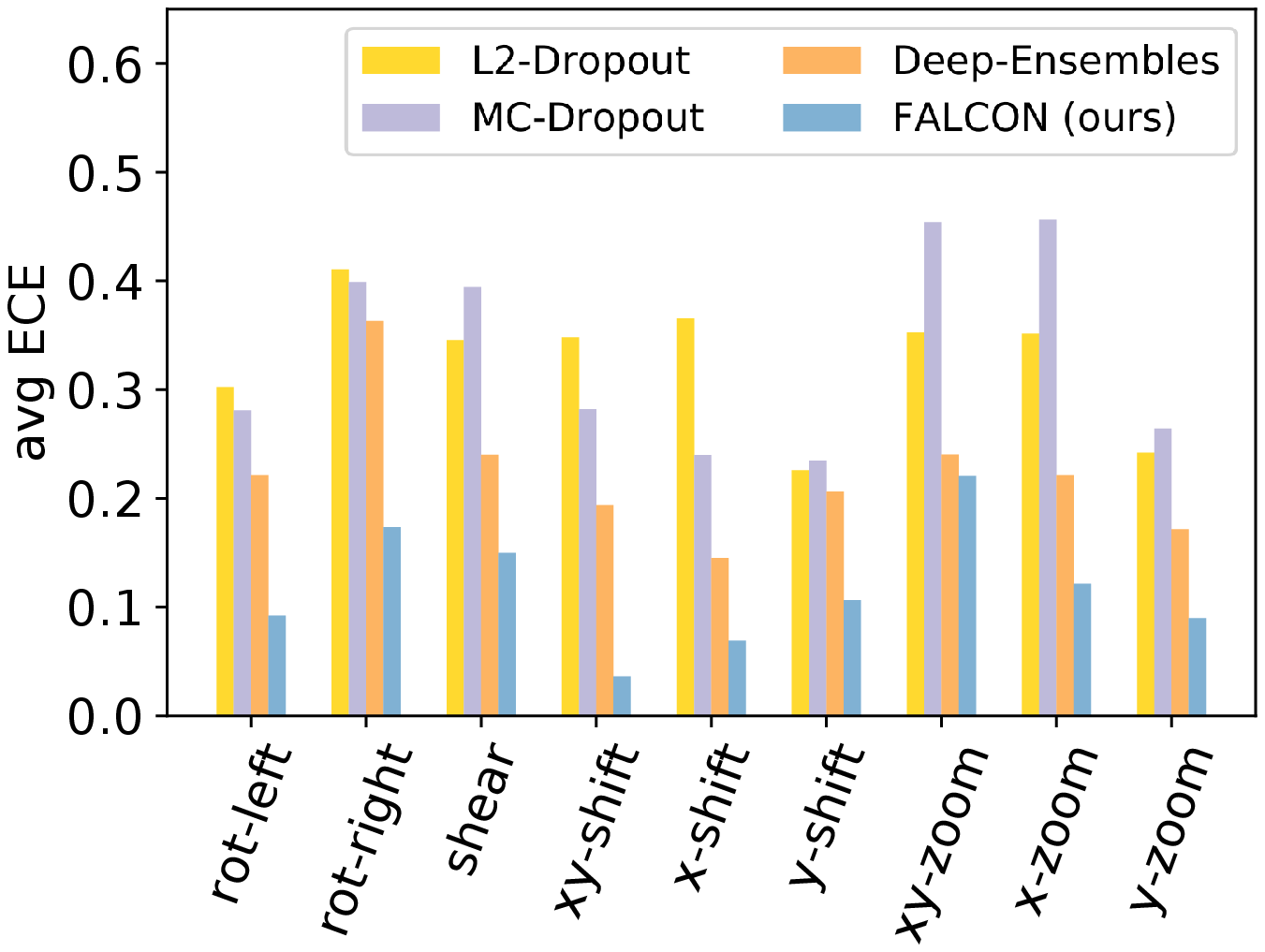}
		\caption{Sequential MNIST - LSTM }
	\end{subfigure}
	\begin{subfigure}[t]{0.27\columnwidth}
		\centering
		\includegraphics[width=\textwidth]{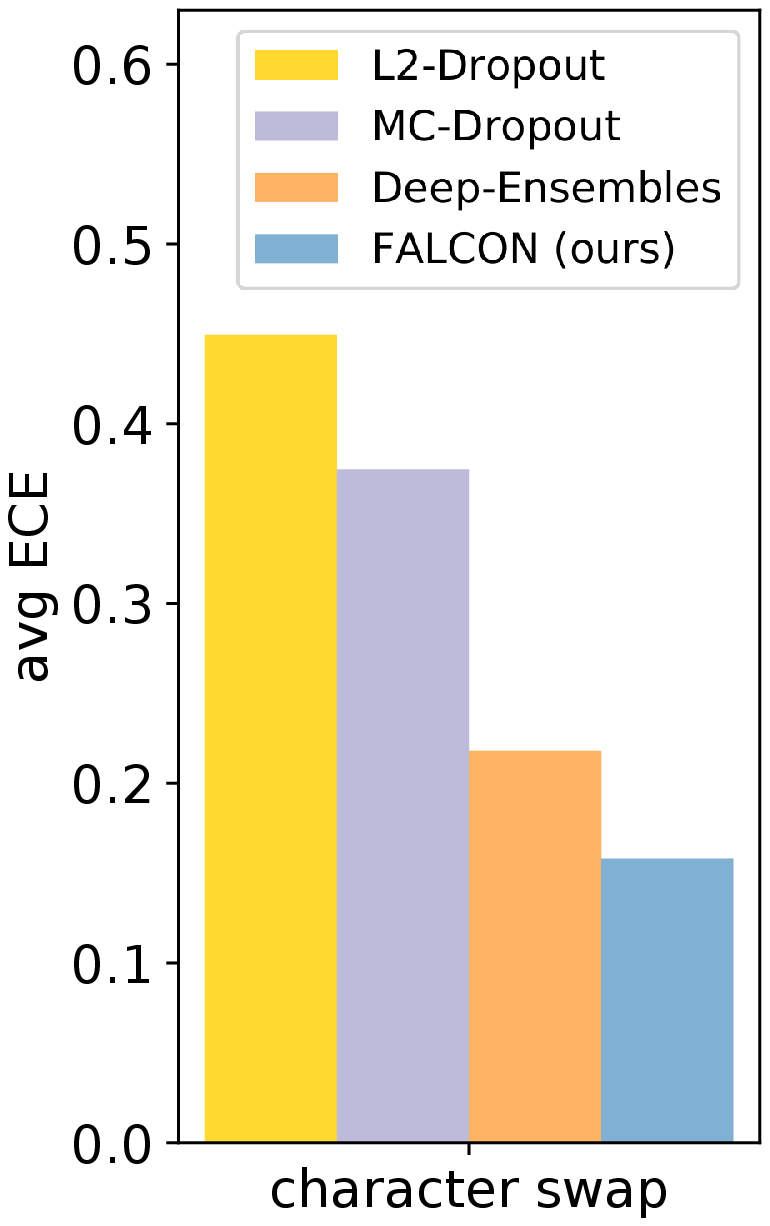}
		\caption{20 Newsgroups}
	\end{subfigure}	
	\caption{Calibration of sequence models for classifying sequential MNIST and 20 Newsgroups data, quantified by computing the micro-averaged expected calibration error (lower is better).}
	\label{fig:seqmod}
\end{figure}

To evaluate our modeling approach for sequence data, we fit models on the Sequential MNIST dataset, where images are converted to pixel-wise sequences of length 28x28. In addition, we fit models on the 20 Newsgroups dataset, where news articles are modelled as sequences of words.\\
We trained recurrent networks with LSTM and GRU cells.  We were not able to train MNF and SVI for the LSTM/GRU models and therefore do not report results for these tasks for SVI and MNF.  \citet{snoek2019can} report a similar issue for SVI and present results for LL-SVI, where Flipout is only used in the last layer; we omit this comparison here since they report that for LSTMs, LL-SVI performs worse than the vanilla method.\\
\textbf{Sequential MNIST}
Fitting LSTM models on sequential MNIST is a challenging task \cite{bai2018empirical}, and it was only possible to achieve state-of-the-art predictive power with EDL for shorter sequences (downsampling of images before conversion to sequence). While performance of GRUs was better for all modelling approaches, EDL also did not achieve a competitive accuracy (0.39 for LSTM and 0.838 for GRU). 
We found that our approach achieved competitive predictive power for LSTM and GRU models (Table \ref{tab:seq_acc}) and substantially improved calibration of the predictive uncertainty for both models (Fig. \ref{fig:seqmod}; results for GRU are reported in Table \ref{tab:seq_acc}). This illustrates that in contrast to existing approaches FALCON is able to yield well-calibrated and trustworthy predictions without compromising on accuracy, even for challenging tasks such as classifying long sequences with LSTMs.\\
\textbf{20 Newsgroups}
To further evaluate the ability of FALCON to model sequence data, we compared the performance of FALCON to existing approaches for an NLP task. To this end, we trained LSTMs to classify news articles into one of 20 classes. We generated vector representations of words using the pre-trained GLOVE embedding (length 100). We trained LSTMs and evaluated trained models on a perturbation strategy based on random word swaps. To establish a perturbation strategy with  gradually increasing perturbations, we varied the fraction of words drawn from each sample between 0\% and 45\% in 5\% steps (gradually decreasing accuracy to random levels).\\
Similar to the LSTM model trained on sequential MNIST, we found that EDL did not achieve competitive predictive power, with an accuracy of 49.3\% only. In contrast, FALCON resulted in well-calibrated  predictions while maintaining a competitive accuracy of 75.7\%, compared to 75.9\%, 72.8\% and 77.3\% for L2-Dropout, MC-Dropout and Deep Ensemble respectively. 
\begin{table}
	\centering
	\caption{Predictive entropy for truly OOD predictions in sequence modelling tasks. sM stands for sequential MNIST.}
	\begin{tabular}{l l l l l}
		\toprule
		&  LSTM-sM & GRU-sM &  20 ng \\ 
		\midrule
		L2-Drp & 0.82 & 0.36 & 0.64  \\ 
		MC-Drp & 0.63 &0.47 &  0.91  \\ 
		DeepEns &  0.86 & 1.06 & 1.82  \\ 
		FALCON & \textbf{1.71} & \textbf{1.61} & \textbf{2.23}  \\ 
		\bottomrule
	\end{tabular}
	\label{tab:ent_seq}
\end{table}
As before, the model confidence of FALCON was substantially better calibrated than existing methods (Figure \ref{fig:seqmod} (b)).\\
\textbf{Predictive uncertainty for truly OOD} For all sequence modelling tasks, we also investigated predictive uncertainty for truly OOD samples. To this end, we computed the mean entropy for truly OOD predictions by considering perturbations at maximum strength (epsilon 90), with model accuracy close to random. In this scenario, the predictive entropy of FALCON was at least 22\% higher compared to baselines, indicating that FALCON is able to make uncertainty-aware predictions even for truly OOD samples (Table \ref{tab:ent_seq}).

\section{Discussion and Conclusion}
We present a fast and generalizable approach for encouraging well-calibrated uncertainty-awareness of deep neural networks. To this end, we combine an entropy encouraging loss-term with an adversarial calibration loss. We show on diverse data modalities and model architectures that our approach yields well-calibrated predictions both for dataset shift and for OOD samples generated based on 32 distinct perturbations and datasets. While SVI and EDL are promising on simple LeNet (trained on MNIST) and VGG19 (trained on CIFAR10) architectures, they are challenging to fit on more complex architectures such as LSTMs and ResNet. Deep Ensembles yield substantial improvements over the L2-dropout baseline, in particular for complex models such as recurrent networks and ResNet. However, training an ensemble of neural networks increases training time linearly with the number of networks in the ensemble. This can be a substantial drawback for applications where training of a single deep network on a large dataset can take several days or even weeks and results in a substantial carbon footprint. \citet{snoek2019can} have recently evaluated calibration under domain drift for a range of existing algorithms and datasets based on ECE. Our findings for baseline methods confirm their results, in particular for deep ensembles and SVI (they did not consider EDL, MNF, VUC and MixUp). Thus, while they also report promising results for SVI on MNIST/CIFAR, they also note that it is difficult to get it to work on larger datasets and other architectures. In addition, they also find that for all baseline methods calibration decreases with increasing dataset shift, resulting in largely overconfident predictions for truly OOD data. In contrast, FALCON outperforms both SVI and EDL on simple architectures (LeNet/VGG19), as well as deep ensembles on complex architectures (LSTM/GRU/ResNet). 
\section{Additional technical details}
\begin{table*}[h]
    \centering
    \caption{Test accuracy and average ECE (lower is better) across all perturbation strategies for LSTM and GRU models. For 20 Newsgroups micro-averaged ECE for character swap is reported. For all models test accuracy is computed on the unperturbed test set.}
    \begin{tabular}{l l l l l l l }
    	\toprule
    	&  \multicolumn{2}{c}{LSTM MNIST} & \multicolumn{2}{c}{GRU MNIST} & \multicolumn{2}{c}{LSTM 20 Newsgroups}\\
    	& Test acc. &  Avg. ECE & Test acc. & Avg. ECE   & Test acc. & ECE\\ 
    	\midrule
    	L2-Dropout & 0.986 & 0.327 & 0.991 & 0.334 & 0.759 &  0.449\\ 
    	MC-Dropout & 0.986 & 0.334 & 0.98 & 0.296 & 0.728 & 0.375 \\
    	Deep-Ensemble & 0.99 & 0.222 & 0.99 & 0.168 & 0.773 & 0.218 \\ 
    	FALCON & 0.978 &  \textbf{0.118} & 0.988 &  \textbf{0.108}  & 0.757 &  \textbf{0.158}\\ 
    	\bottomrule
    \end{tabular}
    \label{tab:seq_acc}
\end{table*}

\section{Parameter and hyperparameter settings}
Deep Ensembles, MNF, and EDL were trained with default values for method-specific hyperparameters (e.g. number of neural networks in a Deep Ensemble). In addition, the following hyperparameters were picked using hyperparameter searches. For all methods, the learning rate was chosen from $\{1e-5, 5e-5, 1e-4, 5e-4, 1e-3, 5e-3\}$. In addition, for the baseline method (L2), our method (FALCON), Deep Ensembles and EDL, dropout was chosen from $\{0, 0.5\}$ and L2-regularisation from $\{0.0, 0.001, 0.005, 0.01, 0.05\}$. For EDL we chose the KL regularisation from $\{0.5, 1., 5., 15., 10., 30., 50., 100. \}$ . For a fair comparison with this state-of-the art model,  we chose $\lambda_S$ from this same set of values for FALCON and $\lambda_{adv}$ from $\{0.25, 1e-1, 1e-2, 1e-3, 1e-4, 1e-6\}$. .\\
For the 20 Newsgroups dataset we used the keras tokenizer to format text samples, converting words into lower case, removing punctuation and special characters \verb|!"#$%&()*+,-./:;<=>?@[\\]^_`{}~\t\n'.|
We used the first 2500 words of an article as input. We trained LSTM models with one hidden layer of 130 hidden units using the RMSPROP optimizer. GRU models were trained with one hidden layer of 250 hidden units to reflect the reduced complexity of GRU cells compared to LSTM cells.

\paragraph{Deep Ensembles}
For deep ensembles of LSTMs trained on sequential MNIST we found that models did not converge when training the networks with standard adversarial examples; we therefore also trained ensembles with a reduced $\epsilon$ of 0.005 and report performance for this modified Deep Ensemble approach. For all other settings, including the deep ensemble of GRUs on sequential MNIST and the deep ensemble of LSTMs on the 20 Newsgroups data, we report performance with standard adversarial training ($\epsilon=0.01$).\\
For all settings except Imagenet we trained a standard ensemble of 5 neural networks. For Imagenet, we used the trained ensemble of 10 neural networks provided as part of \cite{snoek2019can} (which were trained without using adversarial training\footnote{\url{https://github.com/google-research/google-research/tree/master/uq_benchmark_2019/imagenet}}, $\epsilon=0$).
\paragraph{FALCON - Imagenet}
For the Imagenet data, we initialized FALCON with the weights of a pretrained model \cite{snoek2019can} to facilitate faster convergence. We therefore chose a smaller learning rate of 0.000001. For all other experiments we used random initializations.

\subsection{Ablation study and sensitivity analysis} 
\begin{figure}[h]
	\centering
	\includegraphics[width=0.25\textwidth]{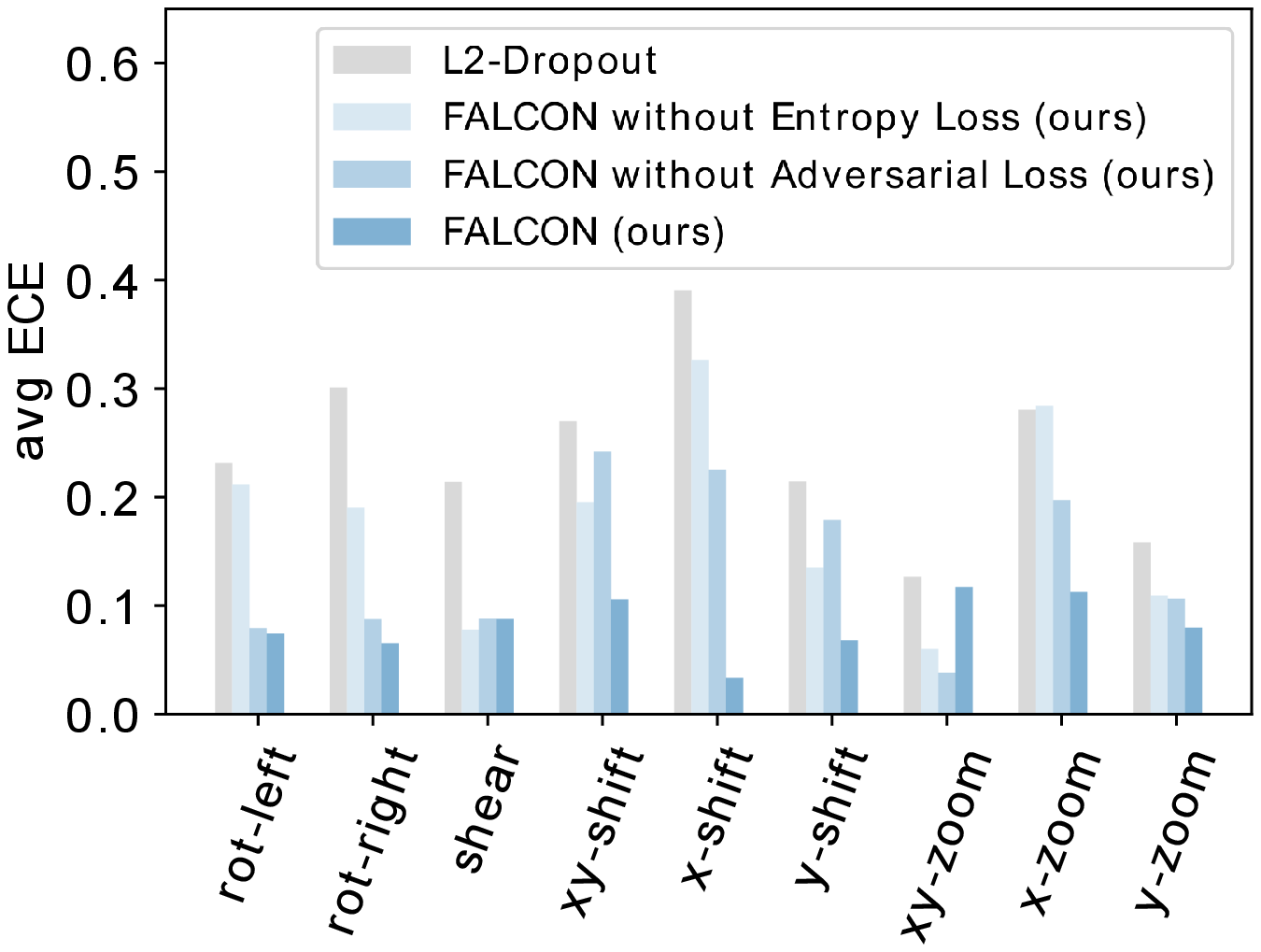}
	\caption{Micro-averaged ECE for FALCON with only one loss term (MNIST).}
	\label{fig:ablation}
\end{figure}

In order to investigate the influence of the individual loss terms on calibration, we performed an ablation study, omitting one of the two loss terms, $L_S$ and $L_{adv}$ respectively. While either loss term results in an improved calibration compared to the L2-dropout baseline, combining both terms yields consistently better results (Figure \ref{fig:ablation}).\\

\begin{figure}[h]
	\centering
	\begin{subfigure}[t]{0.23\textwidth}
		\centering
    	\includegraphics[width=\textwidth]{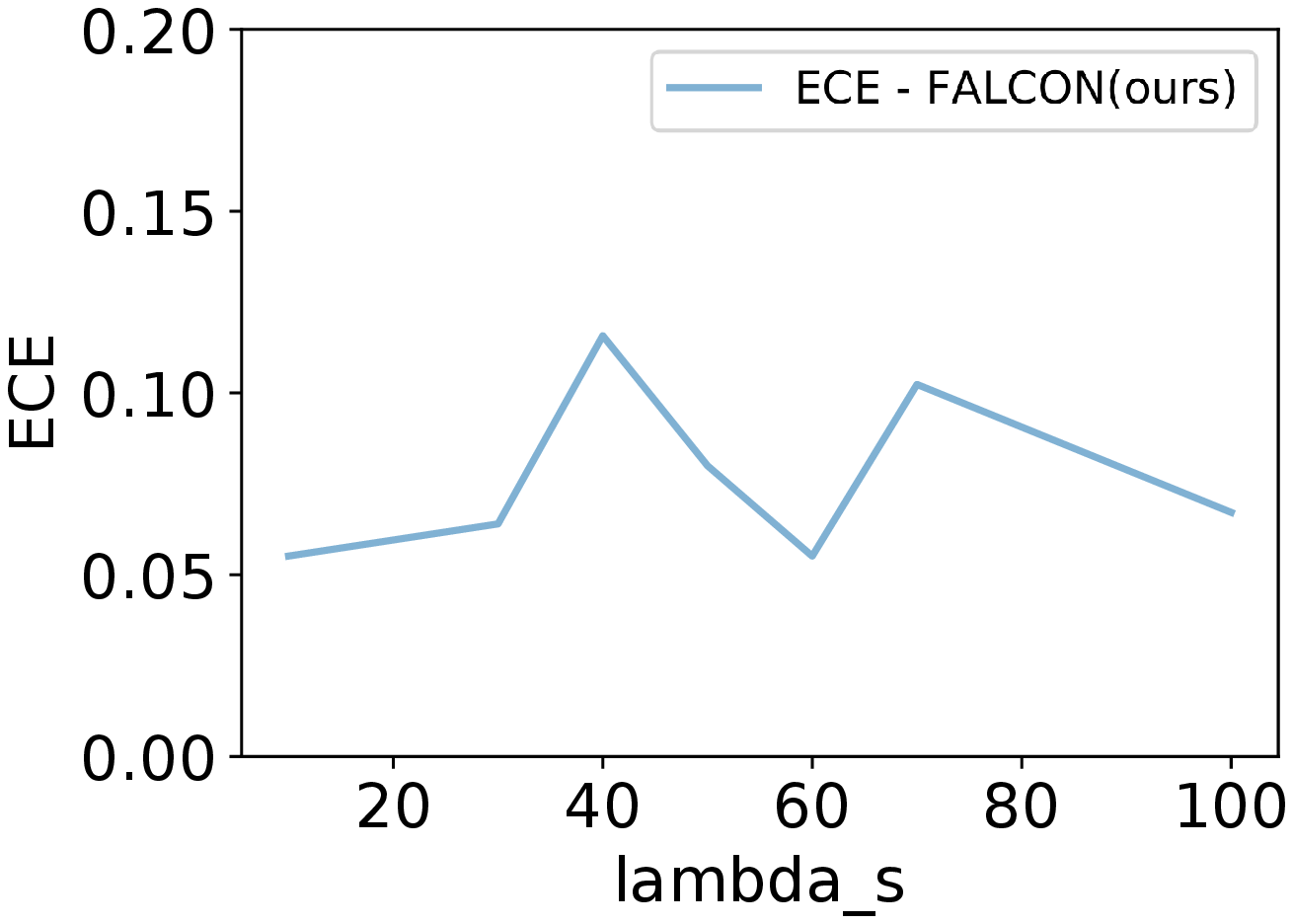}
		\caption{Robustness of $\lambda_S$}
	\end{subfigure}
	\begin{subfigure}[t]{0.23\textwidth}
		\centering
		\includegraphics[width=\textwidth]{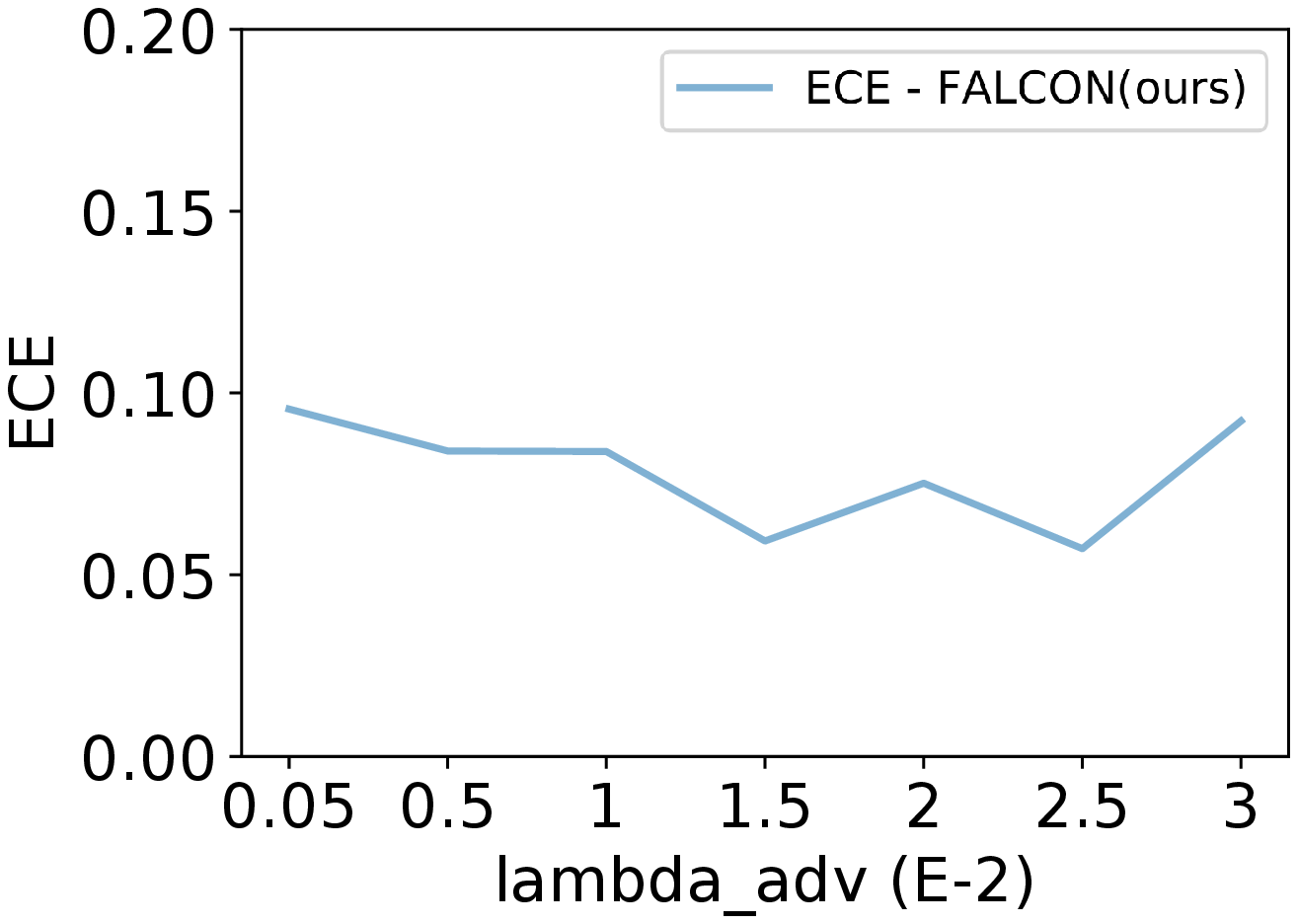}
		\caption{$\lambda_{adv}$ (the x-axis values are scaled by a factor of $10^{-2}$)}
	\end{subfigure}
	\caption{Robustness of hyperparameters, evaluated for the perturbation yzoom on the test set. Even when varying both hyperparameters, micro-averaged ECE for perturbation yzoom was robust.}\label{fig:robust_lambda}
\end{figure}

In addition, we performed a sensitivity analysis, in order to quantify the dependence of model performance on hyperparameter choice. To this end, we trained LeNet models on MNIST varying both hyperparameters over wide ranges. We first fixed $\lambda_S$ at the optimal value of 50 and varied $\lambda_{adv}$ between 0.0005 and 0.03 and computed the micro-averaged ECE for perturbation yzoom for all hyperparameter combinations. Next, we fixed $\lambda_{adv}$ at the optimal value of 0.02  and varied  $\lambda_S$ between 10 and 100. We found that even when varying both hyperparameters over a wide range, ECE remained robust and varied by less than 0.04 for $\lambda_{adv}$ and less than 0.06 for $\lambda_S$ (Fig. \ref{fig:robust_lambda}). Accuracy was not affected by the choice of either $\lambda$ and remained between 0.985 and 0.991.

\section{Supplementary analyses image data}

\subsection{MNIST}

\begin{figure}[h]
	\centering
	\begin{subfigure}[c]{0.2\textwidth}
	\centering
	\includegraphics[width=\textwidth]{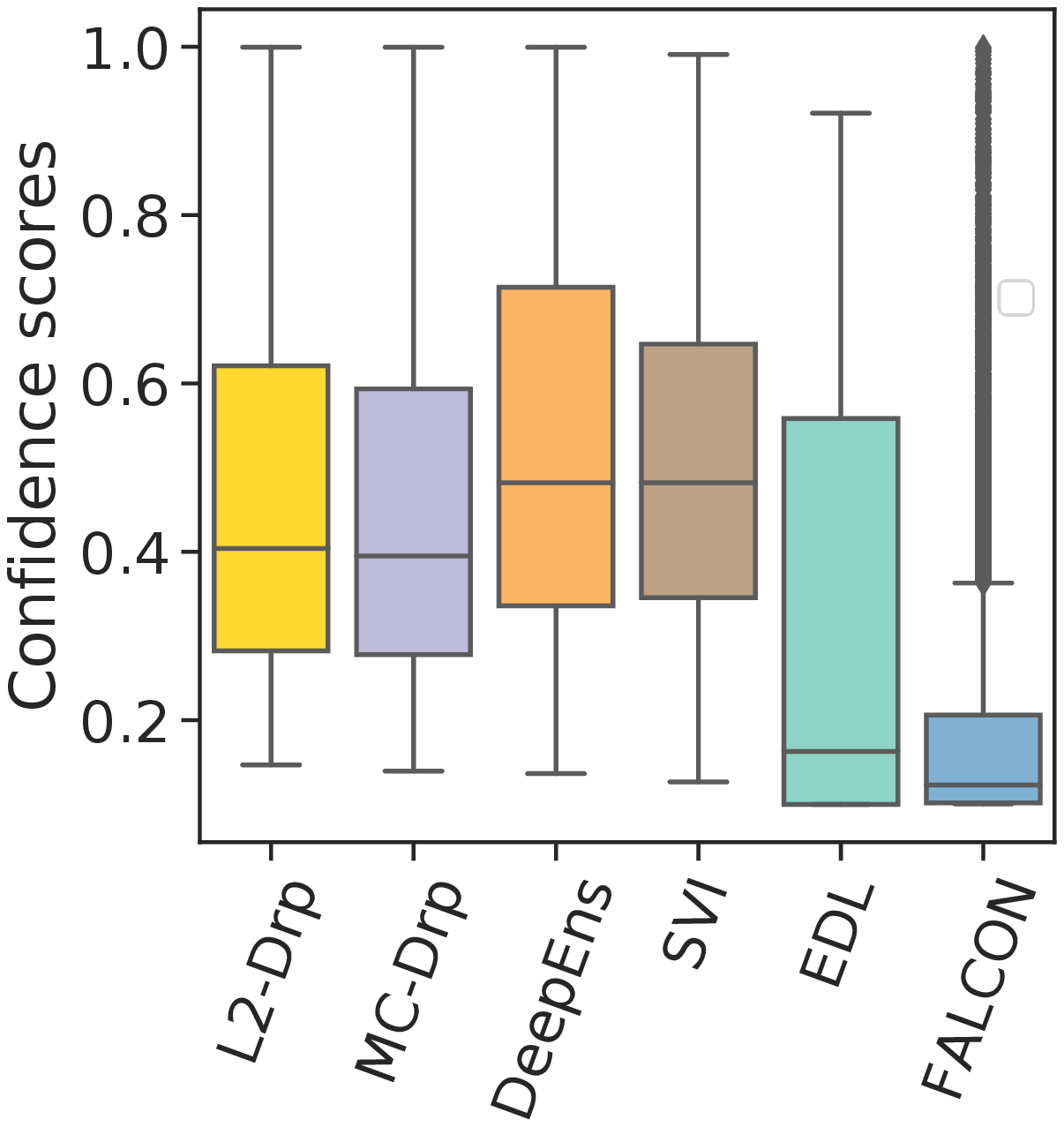}
	\caption{Confidence scores for predictions of Fashion-MNIST.}
	\end{subfigure}
	\begin{subfigure}[c]{0.25\textwidth}
	   \centering
	\includegraphics[width=\textwidth]{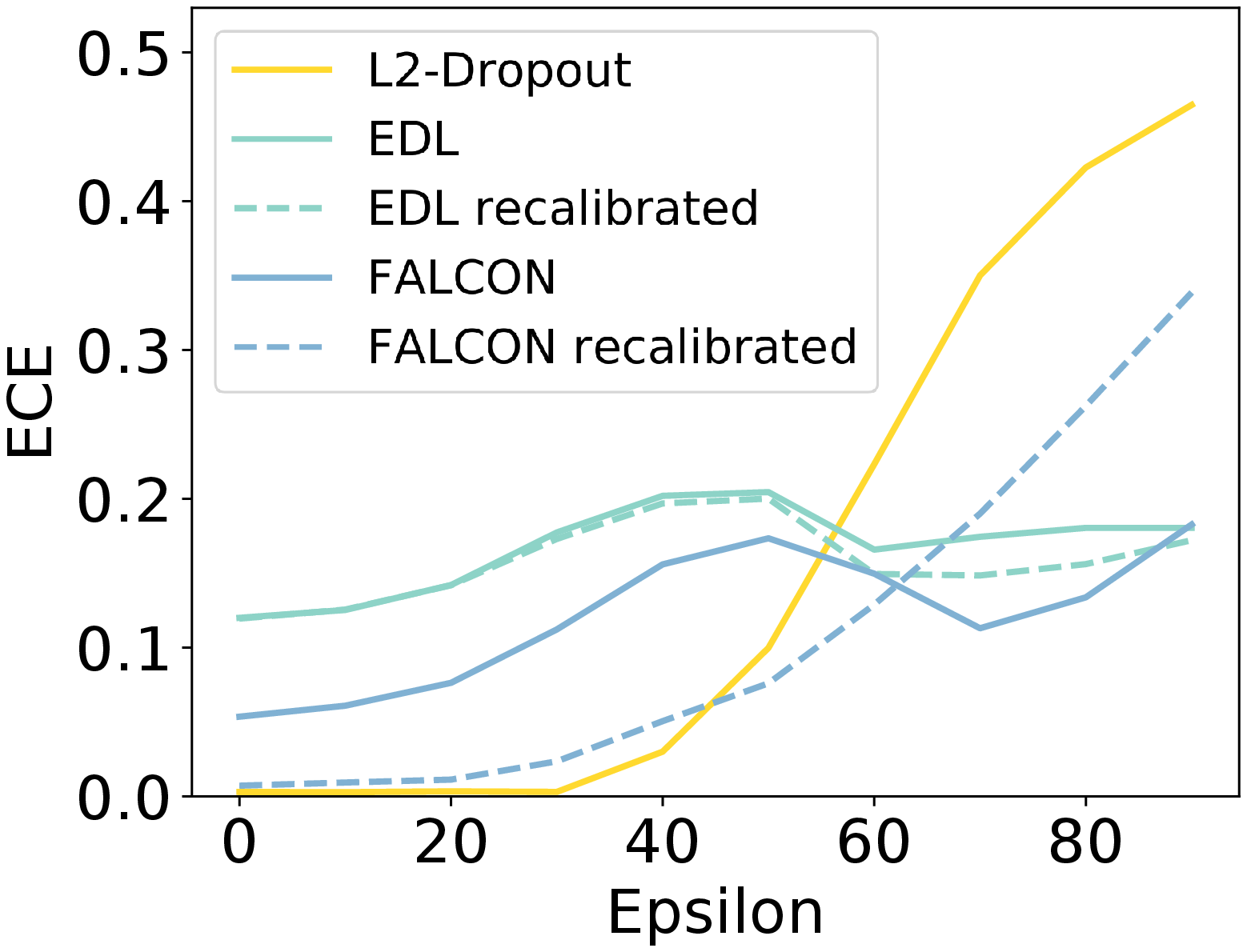}
	\caption{Calibration under increasing levels of y-zoom. EDL recalibrated and FALCON recalibrated refer to models where temperature scaling was applied after training.}	
	\end{subfigure}
	\caption{Additional analyses of MNIST}
	\label{fig:fmnistconf}
\end{figure}

\paragraph{OOD perfromance}
To further analyse the OOD performance of MNIST models on Fashion-MNIST, we assess the distribution of confidence scores when making predictions on this completely OOD dataset (Fig. \ref{fig:fmnistconf}). FALCON has lowest confidence scores, indicating that it is significantly better calibrated and more uncertainty aware than all baselines ($p<10^{-10}$, Wilcoxon signed rank test, FACLON vs EDL). 

\begin{figure}[h]
	\centering
	\includegraphics[width=0.45\textwidth]{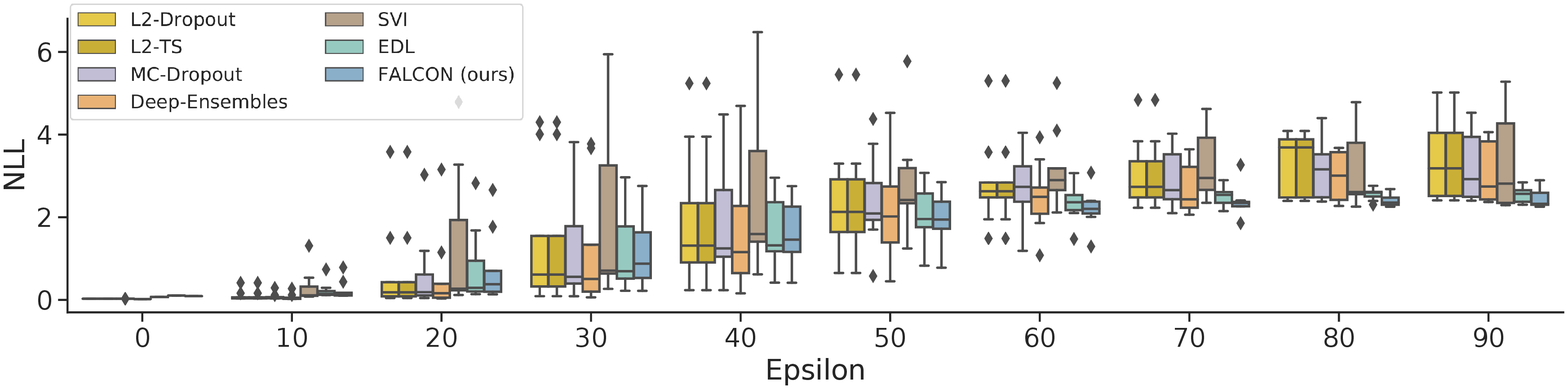}
	\caption{NLL for increasing levels of dataset shift for MNIST}
	\label{fig:nllmnist}
\end{figure}

\paragraph{Negative log likelihood}
 While the focus of our contribution is on the evaluation of the quality of predictive uncertainty (i.e. calibration), it is crucial for any new approach to achieve competitive accuracy, which we show in table 1. In addition, we have computed NLL as proper scoring rule and show in Fig. \ref{fig:nllmnist} that all models achieve comparative NLL for low levels of domain shift, with FALCON and EDL performing significantly better for larger levels of domain shift ($p<10^{-10}$ for epsilon $> 50$, Wilcoxon rank sum test).

\paragraph{Effect of temperature scaling}
Since we observed that FALCON and EDL displayed some underconfidence for in-domain predictions, we investigated whether postprocessing the trained models using temperature scaling would be beneficial. However, we found that for FALCON this resulted in overconfident predictions for samples far away from the training distribution and had little effect on EDL as shown in Fig. \ref{fig:fmnistconf} (b).\\

\begin{figure}[h]
	\centering
				\begin{subfigure}[t]{0.06\textwidth}
		\centering
	\includegraphics[width=\textwidth]{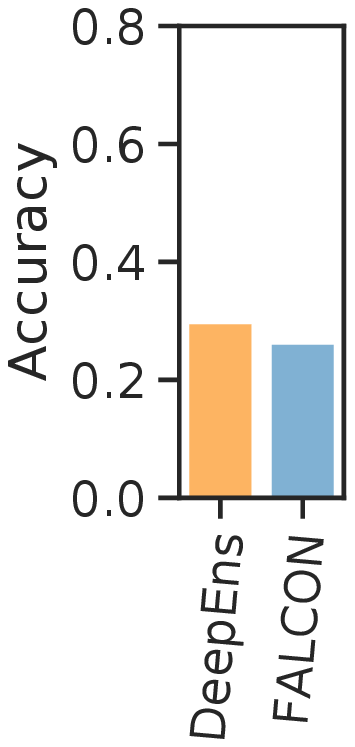}
		\caption{Objectnet}
	\end{subfigure}	
			\begin{subfigure}[t]{0.38\textwidth}
		\centering
		\includegraphics[width=\textwidth]{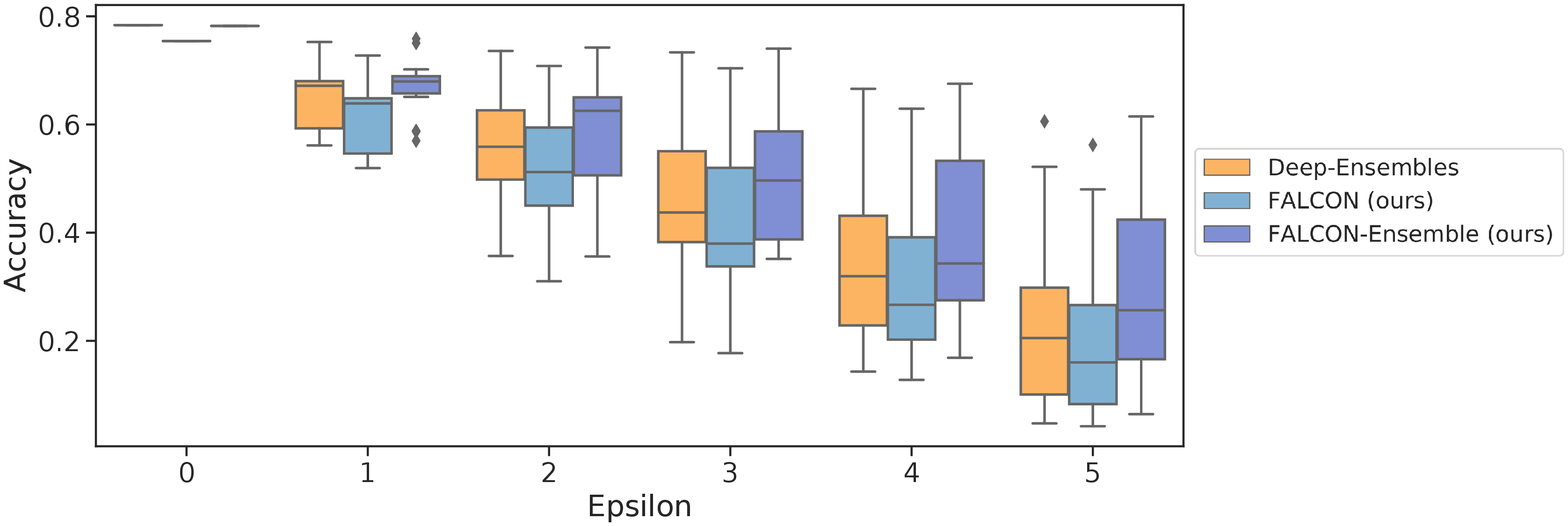}
		\caption{Accuracy for increasing levels of corruption}
	\end{subfigure}
	\caption{Accuracy for FALCON and Deep Ensembles for (a) ObjectNet (overlapping classes) and (b) corrupted ImageNet for increasing levels of corruption.}
	\label{fig:accimobj}
\end{figure}

\begin{figure}[h]

	\centering
		\begin{subfigure}[t]{0.37\textwidth}
	\includegraphics[width=\textwidth]{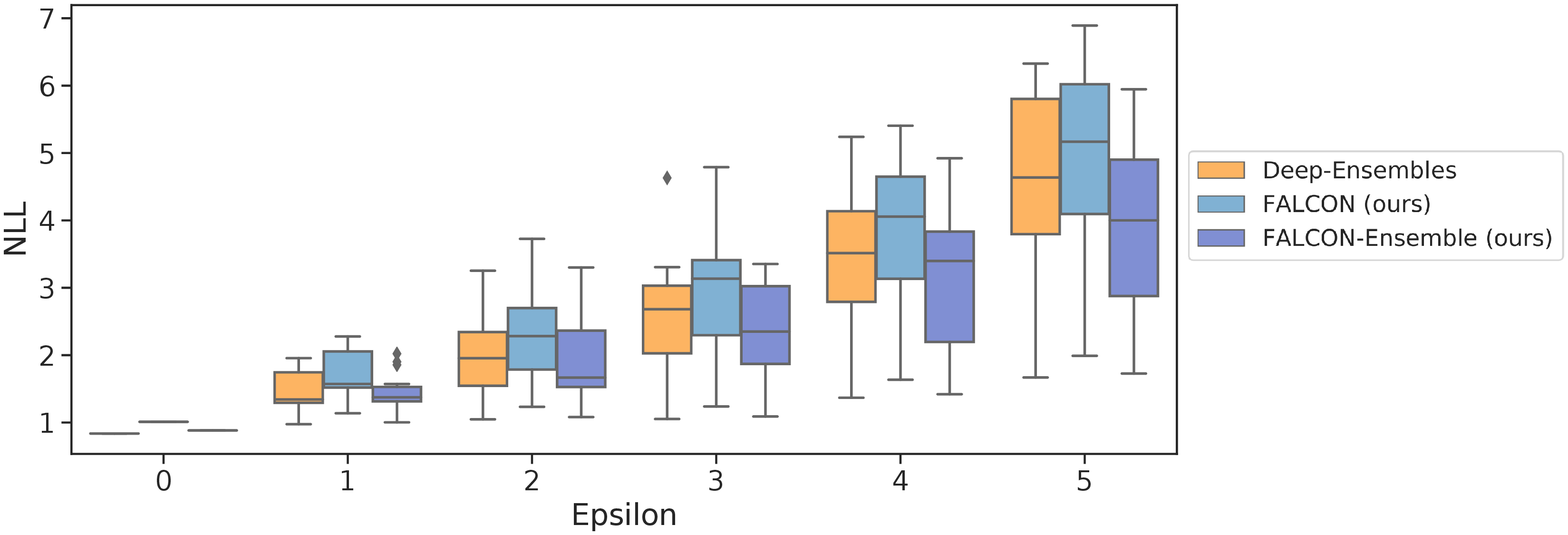}
	\caption{Corrupted Imagenet}
	\end{subfigure}
	\begin{subfigure}[t]{0.08\textwidth}
		\includegraphics[width=\textwidth]{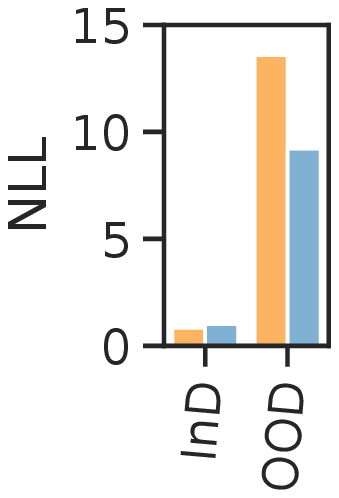}
	\caption{Objectnet}
	\end{subfigure}
    \caption{NLL Imagenet}
	\label{fig:NLLimnet}
\end{figure}

\subsection{ImageNet}

\paragraph{Model accuracy and NLL ImageNet corruptions}
We found that accuracy of both models degraded with increasing corruptions (Fig. \ref{fig:accimobj}). We further observed that levels of accuracy were slightly but consistently higher for Deep Ensembles compared to FALCON. To confirm that this mainly due to the ensemble effect, we also trained an ensemble of FALCON models as described above. Figures \ref{fig:accimobj} and \ref{fig:NLLimnet} illustrate that when we train such ensemble of FALCON models accuracy becomes comparable to levels seen in Deep Ensembles and NLL improves over Deep Ensembles due to improvements in OOD calibration.\\

\section{Ethics statement}
FALCON could be applied in a wide range of use cases where domain shifts after model deployment are expected. This includes a multitude of classification tasks in the physical world, such as predictive maintenance tasks where wear and tear or changing operator behaviour can result in domain shifts. Other relevant machine learning tasks include the analysis of customer data, where changing customer behaviour (e.g. during a recession or health crisis) can lead to a domain shift. Especially safety critical applications where no human supervision is carried out (e.g. autonomous driving cars) would profit substantially from a machine learning model that is aware of its own uncertainty.\\
In this case, our research leads to more transparent predictions. Users know when to trust a model prediction, since calibrated outputs mean that confidence scores broadly match model accuracy for a (set of) prediction(s). In other words, FALCON transparently communicates "how well it knows" via these calibrated confidence scores, for all samples the model may encounter throughout its lifecycle. In case of a low confidence score the user can, depending on the use-case, decide to use fall-back methods or involve a human in the loop. In case of consistently low confidence scores, a retraining of the model can be triggered.\\
Transparent predictions can also help increase fairness. For example, when used in a decision-support context, well calibrated predictions can be used to identify types of input samples that were under-represented during training and for which model accuracy is lower. In this case, low confidence scores would make the inability of the model to make a reliable prediction transparent and wrong decision due to a false trust in the model can be avoided. While we have not explicitly investigated the ability of our modeling approach to identify such biases in the training data and their implications on predictions, we hope that our contribution inspires future research in this direction.\\
Transparent predictions, however, also mean that it is possible to identify OOD samples (or outliers) effectively via their associated low confidence score. This may not always be desired since in certain applications it is a risk to privacy. For example, it may be possible to reveal OOD samples that are related to a protected class such as age, race, or gender and underrepresented in the training data.\\
While our modeling approach yields better calibrated inputs compared to the state-of-the-art, calibration is not perfect and there are therefore still risks related to undue trust in models.

Currently, the best performing model for large data and complex architectures is deep ensembles (section 4 and \cite{snoek2019can}). Our modeling approach is not only better calibrated in domain shift and OOD settings, but also 5-10 times more carbon efficient\footnote{Typically deep ensembles consist of 5 neural networks, in the case of ImageNet/ResNet50, the state-of-the-art presented in \cite{snoek2019can} consists of 10 neural networks. Training resources and carbon footprint increase linearly with the number of nets in the ensemble}. This better sustainability may be an important factor when choosing a modeling approach, especially when networks need to be trained on large data.

\bibliography{references} 

\end{document}